\pdfoutput=1
\documentclass[10pt]{article}
\usepackage[preprint]{tmlr}
\newif\ifanon
\newif\ifarxiv

\usepackage{amsmath,amsfonts,bm}

\def\eqref#1{equation~\ref{#1}}

\def\1{\bm{1}}

\def\vmu{{\bm{\mu}}}
\def\vtheta{{\bm{\theta}}}

\def\vs{{\bm{s}}}

\def\vx{{\bm{x}}}

\DeclareMathAlphabet{\mathsfit}{\encodingdefault}{\sfdefault}{m}{sl}
\SetMathAlphabet{\mathsfit}{bold}{\encodingdefault}{\sfdefault}{bx}{n}

\def\gJ{{\mathcal{J}}}

\def\sR{{\mathbb{R}}}

\newcommand{\E}{\mathbb{E}}

\newcommand{\Cov}{\mathrm{Cov}}

\DeclareMathOperator*{\argmin}{arg\,min}

\usepackage{mathtools}

\newcommand{\vvarepsilon}{{\bm{\varepsilon}}}

\newcommand{\vphi}{{\bm{\phi}}}

\newcommand{\vpsi}{{\bm{\psi}}}

\DeclarePairedDelimiterX{\brackets}[1]{[}{]}{#1}

\newcommand{\Exp}[2][]{\E_{#1}\brackets*{#2}}

\newcommand{\given}{\mid}
\newcommand{\givenbar}{%
  \renewcommand{\given}{\nonscript\;\delimsize\vert\nonscript\;}}

\DeclarePairedDelimiterXPP{\prob}[1]{p}{(}{)}{}{\givenbar#1}
\DeclarePairedDelimiterXPP{\post}[1]{q}{(}{)}{}{\givenbar#1}
\DeclarePairedDelimiterXPP{\probof}[2]{#1}{(}{)}{}{\givenbar#2}

\newcommand{\data}{\vx}              %

\newcommand{\param}{\vtheta}         %
\newcommand{\summary}{\vs}           %
\newcommand{\summarynet}{s_{\vpsi}}  %
\newcommand{\infnet}{q_{\vphi}}      %
\newcommand{\score}{\gJ}             %

\newcommand{\dimsummary}{S}
\newcommand{\dimparam}{D}

\usepackage{graphicx}
\usepackage{hyperref}
\usepackage{url}
\usepackage{booktabs}
\usepackage{multirow}
\usepackage{bbm}

\title{Mitigating Representation Gaps in Amortized Bayesian Inference with Auxiliary Supervision}

\arxivtrue

\renewcommand{\headrulewidth}{0pt}

\author{\name Hans Olischläger \email hi@hans.olischlaeger.com \\  %
      \addr Department of Statistics, TU Dortmund University, Germany
      \AND
      \name Svenja Jedhoff \\
      \addr Department of Statistics, TU Dortmund University, Germany
      \AND
      \name Šimon Kucharský \\
      \addr Department of Statistics, TU Dortmund University, Germany
      \AND
      \name Aayush Mishra \\
      \addr Department of Statistics, TU Dortmund University, Germany
      \AND
      \name Stefan T. Radev \\
      \addr Rensselaer Polytechnic Institute
      \AND
      \name Paul Bürkner \\
      \addr Department of Statistics, TU Dortmund University, Germany
}

\begin{document}

\maketitle

\begin{abstract}
Casting Bayesian inference as a neural network optimization problem targeting an amortized posterior is attractive, as it extends to otherwise intractable statistical models and offers near instantaneous inference for new datasets after prepaying the training cost. Although theory guarantees faithfulness under ideal convergence, practical amortized inference still requires iterating over architectures and optimization choices and ultimately ``satisficing'' under finite simulation, compute, and time budgets. Even the best-performing solution may thus retain avoidable representation gaps that typically require problem-specific fixes. Here, we propose a generic alternative which improves training dynamics with auxiliary guidance losses applied to internal representations. Specifically, we show how such guidance leads to faster convergence when training data is abundant and to better performance when it is scarce. We formalize representation gaps as getting stuck in a local optimum at the information bottleneck between the parts of the network tasked with feature learning and those tasked with conditional distribution learning, and offer a generic diagnostic to separate summary failures from inference failures. Finally, we demonstrate that auxiliary supervision improves convergence speed and accuracy on a range of challenging real-world inference problems.
\end{abstract}

\section{Introduction}

Bayesian inference is a cornerstone of scientific modeling, but it is often computationally intractable for complex, realistic models \citep{cranmer2020frontier}. Amortized Bayesian inference (ABI) addresses this by recasting posterior estimation as a neural network optimization problem \citep{zammit-mangionNeuralMethodsAmortized2025, arruda2025diffusion}: instead of solving inference anew for each dataset, a network learns to approximate the posterior conditional on any data by first training on labeled simulations. Once trained, this network delivers near-instantaneous inference for new data, offering an attractive trade-off between an upfront simulation and training cost and downstream inference speed.

While the theoretical guarantees behind ABI are appealing \citep{frazier2024}, practice looks different. Realistic training runs operate under finite simulation, compute, and time budgets, and practitioners must repeatedly iterate over architectures and optimization strategies. Even the best solution found under these constraints is typically only a "satisficing" one, and can suffer \textit{representation gaps}---systematic discrepancies between the amortized approximation and the analytic posterior \citep{hermansCrisisSimulationBasedInference2022, schmittDetectingModelMisspecification2023, falkiewiczCalibratingNeuralSimulationBased2023a, mishraRobustAmortizedBayesian2026}. In practice, closing such gaps relies on problem-specific fixes, which require a lot of expertise in deep learning and do not transfer easily across applications.

In this work, we propose a generic alternative that improves the training dynamics of ABI directly via auxiliary losses applied to internal representations. We show that this auxiliary supervision yields faster convergence when training data is abundant and improved performance when data is scarce. To understand why such gaps arise in the first place, we formalize them as the network becoming stuck in a local optimum at the information bottleneck separating the components responsible for feature (summary) learning from those responsible for conditional distribution (inference) learning. Building on this formalization, we introduce a generic diagnostic that disentangles summary failures from inference failures, providing a principled way to localize where a representation gap originates.

Taken together, our contributions are threefold: (1) a generic, architecture-agnostic supervision mechanism for improving training dynamics in ABI; (2) a formal account of representation gaps as local optima at the summary–inference bottleneck, along with a diagnostic to distinguish their sources; and (3) an empirical demonstration, across a range of challenging real-world inference problems, that auxiliary supervision improves both convergence speed and final accuracy.

\section{Background}

\subsection{Amortized Bayesian Inference}

An amortized posterior approximation is obtained by minimizing the empirical Bayes risk over a suitably parameterized family of conditional distributions that is amenable to gradient based optimization.
The Bayes risk is the expected value of a strictly proper scoring rule $\score$ that assigns a numerical score based on the predictive distribution $\post{\cdot \given \data}$ and on the value that materializes $\param$.
\begin{align}
    \hat{q} = \argmin_q \Exp[(\param, \data) \,\sim\, \prob{\param, \data}]{\score\big(\post{\cdot \given \data},\, \param\big)} \label{eq:expected_score_general}
    \approx \argmin_q \frac{1}{B}\sum_{b=1}^B \score\big(\post{\cdot \given \data^{(b)}},\, \param^{(b)}\big),
\end{align}
We represent the candidate posterior $q$ by a neural network that is a composition of a summary network $\summarynet$ and an inference network $\infnet$,
\begin{equation}
  \post{\cdot \given \data} = \infnet\big(\cdot \given \summarynet(\data) \big),
  \label{eq:summary-inference-net-composition}
\end{equation}
with summary dimension $\dimsummary = \dim(\summarynet(\data))$ exceeding the parameter dimension $D = \dim(\param)$.
We simply write $\summary$ for the summary vector whenever the underlying data is clear from context.
Both point and distributional estimates of the Bayesian posterior can be obtained from \autoref{eq:expected_score_general} with appropriate choices of $\score$ and parameterizations of $q$.

If $\score$ is strictly proper \citep{gneiting2007}, the analytic posterior is the unique minimizer of \autoref{eq:expected_score_general}, such that perfect convergence implies proper posterior inference.
For $q$ to reach its global optimum $q^\star$, it is also necessary that $\summarynet$ learns maximally informative, and thus \textit{sufficient} \citep{halmos1949} summary statistics $\summarynet^\star(\data)$. Only then do we have
\begin{equation}
  \prob{\param \given \data} = \infnet^\star\big(\param \given \summarynet^\star(\data)\big),
  \label{eq:sufficient-stats}
\end{equation}
which motivates joint optimization of $\summarynet$ and $\infnet$ with respect to \autoref{eq:expected_score_general} \citep{radev2020}.

\subsection{Information bottlenecks and representation gaps}

From an information-theoretic standpoint, the decomposition into $\summarynet$ and $\infnet$ constitutes an information bottleneck \citep{saxe2018} at the learned summary statistics $\summary$.
Consequently, learning summary statistics reduces representation complexity by discarding unnecessary information while maximizing task-relevant information.
On the one hand, such a bottleneck is attractive for its effect of controlling generalization error \citep{kawaguchi2023} and inference speed.
On the other hand, the information bottleneck can exhibit approximation, estimation, and optimization errors \citep{bach2024,rodder2025}.
These arise, respectively, from limited model capacity, finite simulation budgets, and suboptimal training dynamics.
We call the failure to learn maximally informative summary statistics the \textit{representation gap}.

Assume an idealized inference network that returns the exact partial posterior $\infnet^\star(\param \given \summarynet(\data)) = \prob{\param \given \summarynet(\data)}$ conditional on the learned summary statistics $\summarynet(\data)$.
When choosing $\score$ to be the log-score,
the excess over the full-data posterior is a mutual information gap
\begin{equation}
  \Delta(\summarynet)
  = \mathbb{E}_{\data}\Big[\mathrm{KL}\big(\prob{\param \given \data}
      \,\big\|\, \prob{\param \given \summarynet(\data)}\big)\Big]
  = \mathcal{I}(\param; \data) - \mathcal{I}(\param; \summarynet(\data))
  \;\geq\; 0,
  \label{eq:sufficiency-deficit}
\end{equation}
which vanishes if and only if $\summarynet(\data)$ is sufficient.
For a finite-capacity inference network, jointly-optimal convergence of both networks becomes mutually dependent:
Learned summaries need not only be informative, but their representation needs to be usable by the inference network; the inference network itself is pushed to ignore not immediately exploitable summaries.
Conversely, the inference network needs to supply training signal to the summary network in order to improve that representation during training.

\section{Origins and Mitigation of Representation Gaps}

\subsection{Failure to escape fixed points in training dynamics}

We now discuss how the training dynamics of summary--inference network pairs (\autoref{eq:summary-inference-net-composition}) can lead to partial posterior learning. In particular, we show that training may fail to identify informative statistics for all posterior directions, and argue that such suboptimal solutions can be metastable: they are (near-)stationary points that the optimizer escapes only slowly, if at all.

Suppose that, at some point during training, the summaries split into $\summary = (\summary^{(i)}, \summary^{(u)})$, where $\summary^{(i)}$ is informative about a subset of parameters $\param^{(i)}$, while $\summary^{(u)}$ carries (almost) no information about the remaining directions $\param^{(u)}$.
The loss then drives the inference network towards the partial posterior $\infnet(\param \given \summary) \approx p(\param^{(i)} \given \summary^{(i)})\,\prob{\param^{(u)}}$, which falls back to the prior for $\param^{(u)}$ and ignores $\summary^{(u)}$.

This state is self-reinforcing: the inference network ignores $\summary^{(u)}$ because it is uninformative, and $\summary^{(u)}$ stays uninformative because the inference network ignores it.
To see the latter, consider a single training pair $(\param, \data)$ with loss
$\score = -\log \infnet(\param \given \summary)$, where $\summary = \summary_\vpsi(\data)$.
Its gradient with respect to the summary network parameters $\vpsi$ is
 \begin{align}
  \frac{\partial \score}{\partial \vpsi}
  =
  - \frac{1}{\infnet}
  \frac{\partial \infnet}{\partial \summary}
  \frac{\partial \summary}{\partial \vpsi}
  =
  - \frac{1}{\infnet}
  \left[
    \frac{\partial \infnet}{\partial \summary^{(i)}}
    \frac{\partial \summary^{(i)}}{\partial \vpsi}
    +
    \frac{\partial \infnet}{\partial \summary^{(u)}}
    \frac{\partial \summary^{(u)}}{\partial \vpsi}
  \right].
  \label{eq:gradient-decomposition}
\end{align}
The only term that could make $\summary^{(u)}$ informative is proportional to
$\partial \infnet / \partial \summary^{(u)}$, which vanishes when the inference network ignores $\summary^{(u)}$.
The partial posterior is therefore a saddle point \citep{saxe2014exact}, which gradient descent escapes only slowly.
In stochastic training, this weak escape signal on the weights $\vpsi$ is easily masked by minibatch noise from the dominant first term, which keeps trying to refine $\summary^{(i)}$.

\subsection{Diagnosing unresolved posterior directions}
\label{sec:canoncial_corr_meth}

To diagnose how much information about the parameters is contained in the summaries, we propose a diagnostic based on canonical correlation analysis \citep[CCA;][]{hotelling1936relations}.
Concretely, for summary vector $\mathbf{s} \in \mathbb{R}^{S}$ and parameter vector $\boldsymbol{\theta} \in \mathbb{R}^{D}$, the goal is to quantify whether $\mathbf{s}$ preserves the directions of variation contained in $\boldsymbol{\theta}$. In particular, we ask: \textit{Can every relevant linear combination of the target variables be recovered from some linear combination of the learned summaries?}

CCA first whitens both summary and parameter vectors individually by computing
    $\tilde{\mathbf{s}}  = \boldsymbol{\Sigma}_{s}^{-1/2}\mathbf{s}$ and
    $\tilde{\boldsymbol{\theta}} = \boldsymbol{\Sigma}_{\theta}^{-1/2}\boldsymbol{\theta}$, such that $\operatorname{Cov}(\tilde{\mathbf{s}})=\mathbf{I}$ and $ \operatorname{Cov}(\tilde{\boldsymbol{\theta}})=\mathbf{I}$.
Their cross-covariance is then
\begin{equation}
    \operatorname{Cov}(\tilde{\mathbf{s}},\tilde{\boldsymbol{\theta}})
    =
    \operatorname{Cov}(\mathbf{s})^{-1/2}
    \operatorname{Cov}(\mathbf{s},\boldsymbol{\theta})
    \operatorname{Cov}(\boldsymbol{\theta})^{-1/2}.
\end{equation}
Its singular values $\rho \in [0, 1]^R$, with $R = \min(S, D)$, are known as \textit{canonical correlations}, which can be sorted as $1 \geq \rho_1 \geq \ldots \geq \rho_R \geq 0$. Since the canonical directions are mutually uncorrelated, CCA finds the coordinate systems in which summaries and parameters are maximally aligned.

If $S \geq D$, the standard in ABI, the canonical correlation $\rho_i$ describes how well the $i$th independent direction in parameter space is represented by the summary space. To diagnose how well the whole parameter space is represented, we propose to use $\rho_{\min} = \rho_R$ as a diagnostic since it measures the \emph{worst-preserved parameter direction}.
When $\rho_{\min} \approx 1$ against a posterior-mean target, all parameter directions are almost perfectly represented by the summaries.
Conversely, a small $\rho_{\min}$ indicates that at least one parameter direction is absent or only weakly represented by the learned summaries. Alternatively, using the mean correlation can be a useful diagnostic too but may hide failures in recovering individual parameters.

CCA requires access to parameter-summary pairs. In simulated scenarios, ground-truth parameters can be used. In this case, the CCA diagnostic indicates the informativeness of the summaries to recover the ground truth, but $\rho_{\min}$ is then bounded by $\sqrt{1-v}$, with $v$ the largest normalized expected posterior variance over parameter directions, and should be read against that ceiling (Appendix~\ref{sec:canoncial_corr_app}).
Alternatively one can use a point estimate (e.g., posterior mean) obtained from a reference approach (e.g., MCMC). In that case, CCA indicates the informativeness of the summaries to recover that posterior point. We apply the latter in our case studies, since it also works for empirical data without ground-truths.

\subsection{Guidance losses in summary space}
\label{sec:guidance_method}

Our proposed losses guide summary dimensions towards learning posterior point estimates of the parameters.
Consider $K$ point estimates of interest (e.g., posterior mean or quantiles) towards which the summary should be guided. For simplicity, we assume that guidance is applied to all $D$ parameters. Then, we need a summary dimension $S \geq KD$. To ensure sufficient flexibility of the summary space,
we recommend $S \geq (K+1)D$, such that at least $D$ summary dimensions are unaffected by the guidance losses. Let $\ell_{k}$ be the loss function to learn the $k$th point estimate \citep[e.g., the squared loss for the posterior mean;][]{gneiting2007} with a weight $\lambda_k$ to control its influence on the overall loss. Further, we denote the specific summary dimension  learning the $k$th point estimate for the $d$th parameter as $\summary_{kd}$. The proposed joint loss is then given by
\begin{equation}
\hat{q},\hat{s} = \argmin_{q,s} \Exp[(\param, \data) \,\sim\, \prob{\param, \data}]{\score\big(\post{\cdot \given \summary(\data)},\, \param\big) +
  \sum_{k=1}^K \sum_{d=1}^D \lambda_k \, \ell_k \left(\summary_{kd}(\data), \param_d \right)}.
  \label{eq:guidance_loss_summary}
\end{equation}
Recommendations for practical choices of $\lambda_k$ are provided in Appendix \ref{sec:choose-weight}.

We additionally propose a second variant of our guidance approach, which imposes the losses not on the summary space directly, but on a set of learnable functions $f = (f_{1}, \ldots, f_K)$, conditional on the summaries. Each $f_k$ has output dimension $D$ to learn the $k$th point estimate for all $D$ parameters:
\begin{equation}
\hat{q},\hat{s},\hat{f} = \argmin_{q,s,f} \Exp[(\param, \data) \,\sim\, \prob{\param, \data}]{\score\big(\post{\cdot \given \summary(\data)},\, \param\big) +
  \sum_{k=1}^K \sum_{d=1}^D \lambda_k \, \ell_k \left(f_k(\summary(\data))_d, \param_d \right)},
  \label{eq:guidance_loss_ensemble}
\end{equation}
The functions $f_k$ can be standard MLPs. Together with $q$, they form an ensemble of $K+1$ members sharing the same summary network. The advantage of the ensemble approach is that the summaries are not directly restricted by guidance at the expense of slightly increasing architectural complexity.

While our guidance losses are defined for arbitrary point estimates, we focus primarily on the posterior mean in our experiments, for both theoretical and empirical reasons. Theoretically,
for posteriors of conjugate exponential family models with known sample size, the posterior mean $\bar{\param}(\data)$ is already a sufficient statistic, that is, $\prob{\param \given \data} = \prob{\param \given \bar{\param}(\data)}$ \citep{diaconisConjugatePriorsExponential1979a}. Outside of the exponential family, this is not guaranteed \citep{chen2021neural}; indeed, used alone as a summary statistic in approximate Bayesian computation (ABC), the posterior mean performs clearly worse than standard neural ABI \citep{arruda2026b}. Nonetheless, posterior means remain ``weakly'' sufficient in the sense that they retain first-moment information \citep{fearnhead2012}.
Empirically, we see these theoretical results confirmed in that guidance by the posterior mean alone already fixes the existing representation gaps in all but one case study (see Section \ref{sec:experiments}).

\section{Related work}

\citet{sainsbury-dale2024} formalize neural point estimators using estimate-specific loss functions. Posterior mean estimates have also been used as summary statistics for ABC \citep{fearnhead2012, jiang2017learning}. We build on this line of work, but use point estimates to guide training of a full posterior network, rather than as standalone estimates or as summary statistics for ABC.

Our approach guides standard ABI training with an auxiliary loss, counteracting problematic training dynamics. Previous work has introduced auxiliary losses which can have similar effects.
\citet{falkiewiczCalibratingNeuralSimulationBased2023a} add a differentiable relaxation of the expected coverage error. Balancing losses \citep{delaunoy2022towards,delaunoy2023balancing} bias ratio and posterior estimators toward conservative posteriors.
Both losses mitigate overconfident posteriors. However, calibration and balance are necessary but not sufficient conditions for a correct posterior approximation: the prior itself is perfectly calibrated and balanced. These losses therefore cannot distinguish a well-calibrated posterior approximation from one that ignores the data and returns the prior. In contrast, our training guidance fixes the representation gap that can lead to exactly this failure.

Methods from unsupervised domain adaptation \citep{elsemueller2025uda,huang2023learningRobustStatistics,swierc2024domain,khoo2026minimum} add a domain-alignment loss that matches the distribution of summary statistics between simulated and empirical data, rendering the resulting networks more robust to distribution shifts. \citet{elsemueller2025uda} show that the domain-alignment changes the inference target from the exact posterior to a posterior on adjusted data. Our guidance does not alter the inference target, but merely improves the training dynamics.

\section{Experiments}
\label{sec:experiments}

In the following, we demonstrate empirically how our guidance losses improve neural ABI on one toy and four real-world case studies. All experiments follow the same general setup described below.

\paragraph{General setup.} Every compared estimator within an experiment shares the same summary network configuration so that the performance differences can be attributed to the additional training signal via the auxiliary losses rather than the architecture or capacity. We compare four methods: First, our baseline is a neural posterior estimator consisting of a summary network with a conditional generative inference network, which is either coupling flow \citep{dinh2016density,ardizzone2018analyzing} or flow-matching \cite{lipman2022flow}, trained with Adam \citep{kingma2014adam} and a cosine learning rate decay. In this baseline, the summary network receives gradients only through the inference loss. Second and third, our proposed guidance methods, which add an auxiliary point-estimate loss either on the summary space directly, or in the form of an ensemble with separate point estimate heads (see Section \ref{sec:guidance_method}). In both cases, the auxiliary loss is weighted so that its contribution is comparable in magnitude to the standard loss throughout training, ensuring neither objective dominates the shared representation.
Fourth, a point-estimate network with posterior mean score as training objective as a baseline reflecting point-only estimation. All networks were implemented in \texttt{BayesFlow} \citep{kuhmichel2026}.

\paragraph{Metrics.}
We evaluate each trained approximator on $M$ freshly simulated test instances $\{(\data^{(m)}, \param^{(m)})\}_{m=1}^{M}$, disjoint from the data used during training, drawing $S$ posterior samples $\{\hat\param^{(m,s)}\}_{s=1}^{S} \sim \post{\cdot \given \data^{(m)}}$ from each approximator. We report three parameter-specific metrics, computed per parameter component $j = 1, \dots, \dimparam$ and, where applicable, aggregated across the $M$ test instances (see Appendix~\ref{sec:metrics_app} for details): The \textit{RMSE} of posterior mean estimate vs. ground-truth measures parameter recovery. \textit{Posterior contraction} reflects how much the posterior narrows relative to the prior. \textit{Calibration error} measures whether credible intervals achieve their nominal coverage. As two global, parameter-independent metrics, we report the \textit{expected log posterior density} --- the log-probability the trained neural posterior estimate assigns to the true parameters given the test data --- and the \textit{CCA diagnostic} (see Section~\ref{sec:canoncial_corr_meth}), which quantifies how much parameter information the summary network retains.
For the neural point estimation networks, no posterior samples are available, so only RMSE and CCA can be evaluated in this case.

\subsection{Experiment 1: Simple location-scale models}

\begin{figure}[ht]
    \centering
    \includegraphics[width=\linewidth]{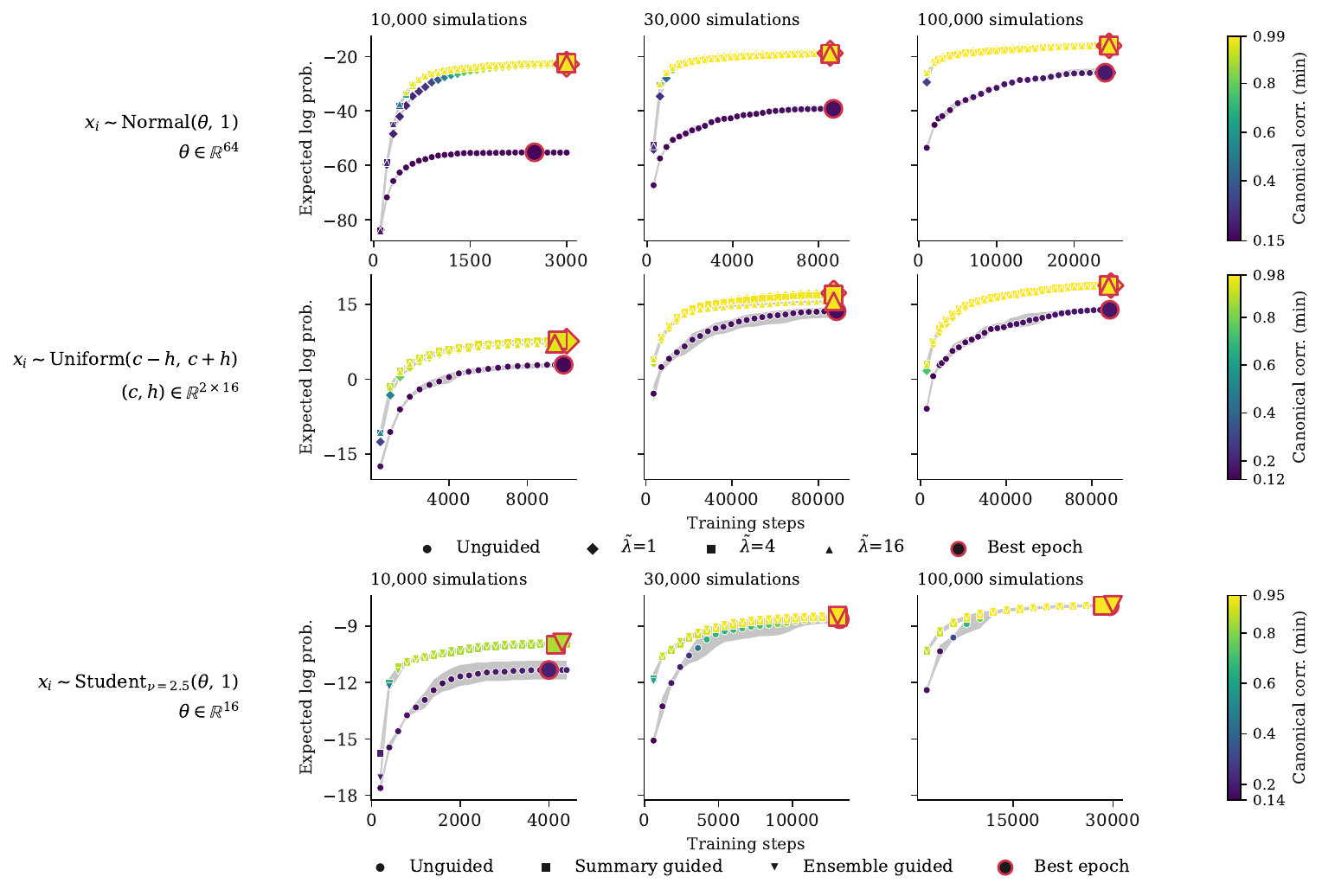}
    \caption{\emph{Location-scale models (Experiment 1)}. Trajectories of posterior accuracy, measured by the expected log probability (y-axes), as well as summary informativeness, measured by the minimum canonical correlation (color) during training.
      The effectiveness of our guidance losses is evident in both metrics and across inference tasks, with parameter dimension $D=64, 32$ and $16$ for the  \textit{Normal}, \textit{Uniform} and  \textit{Student-t} likelihood respectively (rows).
      All models are trained until convergence but with simulation budgets varying from 10k to 100k (columns).
      For \textit{Normal} and \textit{Uniform} tasks guidance is applied directly to half of the summary statistics with varying weight factor $\tilde \lambda$. For \textit{Student-t}, we additionally include ensemble guidance, fixing $\tilde \lambda=4$ for both guidance losses.
  }
    \label{fig:experiment-1}
\end{figure}

Multivariate location-scale distributions allow us to study the effect of point estimate guidance for summary learning as a function of parameter and condition dimensionality, while using highly accurate reference posteriors (either analytic or grid quadrature).
We use standard normal priors throughout and consider different likelihoods,
each producing \mbox{$n=10$} i.i.d. observations $x_i, i \in \{1, \dots n\}$ from a different location-scale distribution corresponding to different sufficient statistics:
\mbox{(1) A \textit{Normal} distribution} with fixed scale, for which we learn the mean parameter $\theta \in \sR^D$. By virtue of being in the exponential family, the sample mean, or equivalently the posterior mean of $\theta$, is minimally sufficient (see Section \ref{sec:guidance_method}). Posteriors are analytic and normal themselves.
\mbox{(2) A \textit{Uniform} distribution} over $\sR^{D/2}$, parameterized by a center and a half-width per dimension, such that  $\theta \in \sR^{D}$. Uniform is not in the exponential family, but still permits finite dimensional sufficient statistics, namely the minimum and maximum value in $\data$ per dimension. Posteriors are analytic with sharp, edge-like boundaries.
\mbox{(3) A \textit{Student-t} distribution} with fixed scale and low degrees of freedom, for which we learn the mean parameter $\theta \in \sR^D$. No sufficient statistic smaller than $\data$ exists. Instead the summary network has to learn an informative compression of the full order statistic. Posteriors are non-analytic with heavy tails.
Details on architectures and training setup are in Appendix \ref{sec:location_scale_app}.

\paragraph{Results.} \autoref{fig:experiment-1}
shows that our loss guidance approaches speed up learning of informative data summaries significantly in all investigated cases.
For both \textit{Normal} and \textit{Uniform}, unguided inference exhibits a representation gap that does not vanish even for 100k training simulations. Specifically, entire parameter directions remain unlearned as indicated by the CCA diagnostic; an issue that our guidance losses fully resolve.
For \textit{Normal}, the posterior mean is minimally sufficient and posterior-mean guidance can quickly and fully resolve the representation gap already for small simulation budgets.
For \textit{Uniform}, optimal posterior inference requires more than just the mean, but rather sample minimums and maximums.
Nonetheless, the representation gap is closed by training with posterior mean guidance as indicated by the CCA diagnostic.
For \textit{Student-t},  no posterior point estimate is sufficient on its own.
Yet, posterior-mean guidance remains highly beneficial: it fully resolves the representation gap already for small simulation budgets and also yields more stable training dynamics (less variation between training seeds). Conversely, for unguided training, resolving the representation gap requires a significantly higher simulation budget and more training steps.

\subsection{Experiment 2: Human decision making}

\begin{figure}
    \centering
    \includegraphics[width=\linewidth]{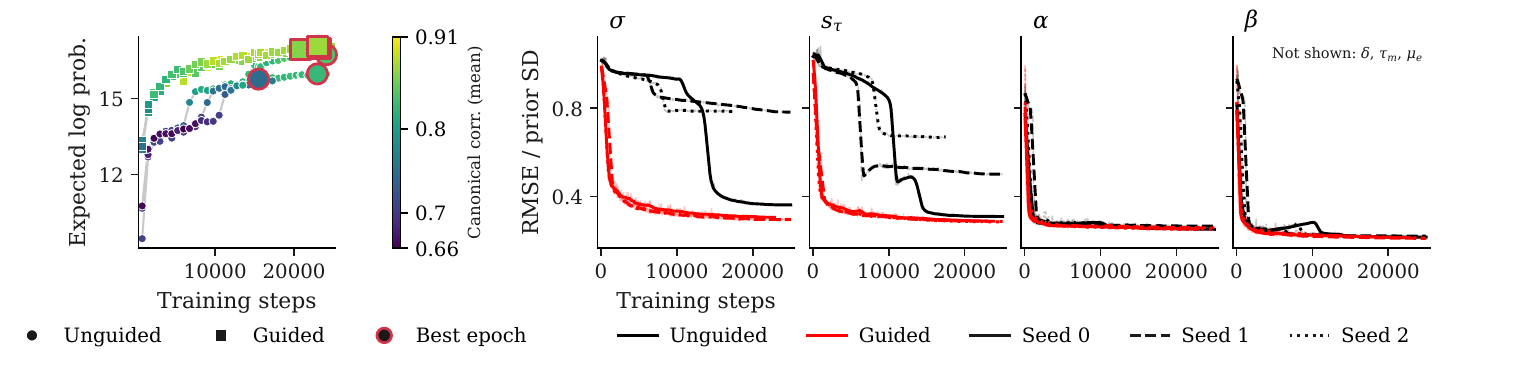}
    \caption{\emph{Human decision making (Experiment 2)}. Guidance removes a long plateau in parameter recovery on the joint EEG--drift-diffusion model at a 10k simulation budget.
        Overall posterior accuracy and summary informativeness (left), posterior-mean error on the prior scale for the four worst-recovered parameters (right).
        Unguided runs (black) leave $\sigma$ and $s_\tau$ near their prior spread for thousands of steps, each seed (line style) breaking at a different point, while guided runs ($\tilde \lambda=4$, red) recover both immediately and without seed spread.
    }
    \label{fig:experiment-2}
\end{figure}

As a first real-world inference task, we consider model 1c of \citet{ghaderi-kangavari2023},
a joint drift-diffusion model of decision making, response time, and single-trial EEG, connecting electrophysiological measurements with human behavior.
A latent, trial-wise encoding time adds to the non-decision time and is observed through the N200 latency, an EEG marker of visual encoding time.
Each dataset has 120 trials of (response time, choice, N200 latency), and the model has seven parameters with uniform priors. See Appendix \ref{sec:ddm-app} for more details and additional results.

\paragraph{Results.} \autoref{fig:experiment-2}
shows how our guidance losses mitigate flawed training dynamics leading to plateaus in parameter recovery.
Specifically, without guidance, the parameters that are identified the weakest are learned late, if at all. According to the CCA diagnostic, we can attribute this to a representation gap:
Improvements in overall posterior accuracy (expected log probability) coincide with a reduction in error of the posterior means for $\sigma$ and $s_\tau$ specifically.

\subsection{Experiment 3: Eye movements}

We investigate the benefits of the proposed methods for inferring parameters of a computational model of eye movement control in scene viewing \citep{nuthmann2010crisp,walshe2021computational}. The model represents fixation durations as a sequence of autonomous timer, labile, and non-labile saccadic programming stages, each an independent discrete random walk with a common threshold $\alpha$ but distinct transition rates ($\tau_\text{timer}$, $\tau_\text{labile}$, and $\tau_\text{nonlabile}$). This setup exhibits a partial information bottleneck: two parameters ($\tau_\text{timer}$, $\tau_\text{labile}$) are recovered with unguided training, whereas parameters $\alpha$ and $\tau_\text{nonlabile}$ are recovered poorly.

\begin{figure}
    \centering
    \includegraphics[width=\linewidth]{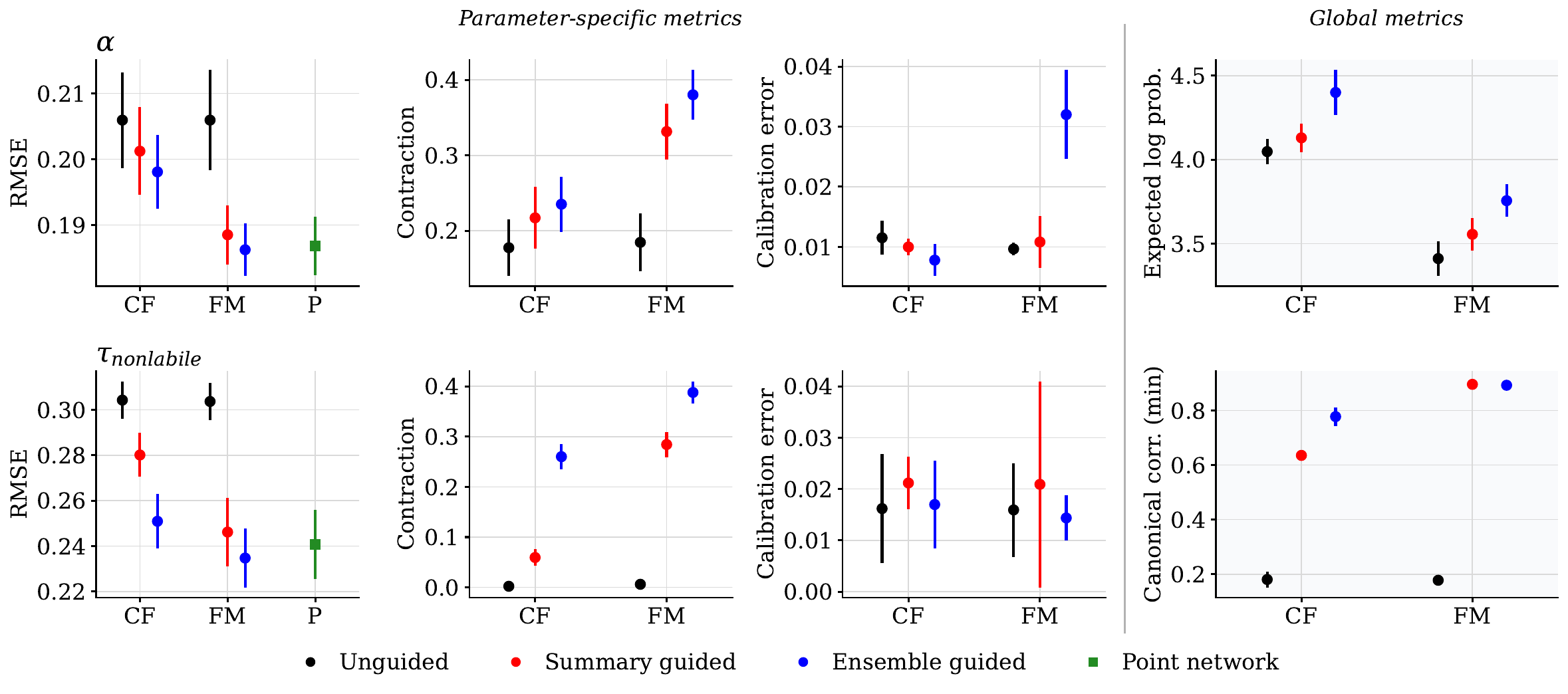}
    \caption{\emph{Eye Movements (Experiment 3)}. Parameter-specific metrics for two parameters, and global metrics. Metrics averaged over three independent test sets of 500 simulations each. Error bars represent standard deviation across the test sets. CF = Coupling flow, FM = Flow matching, P = Point estimation network (posterior mean).}
    \label{fig:crisp-main}
\end{figure}

\paragraph{Results.} \autoref{fig:crisp-main}
shows metrics for the two parameters that are difficult to recover without guidance as well as global metrics. Here, we show results with a set transformer as summary network. Full results over all parameters and experimental configurations are reported in Appendix~\ref{sec:crisp_app}. Both summary and ensemble guided training outperform unguided training. Ensemble guided training gives the best RMSE and contraction; summary guided training improves these metrics for flow matching inference networks but not as much for coupling flows. Calibration error stays relatively stable. The overall improvement is clearly shown in terms of higher expected log probability as well as improving the minimum canonical correlation.

\subsection{Experiment 4: Strong gravitational lensing}
\label{sec:exp_lens}

We demonstrate the efficacy of our proposed methods in inferring strong gravitational lensing parameters from simulated lensing images. Strong lensing occurs when a sufficiently dense foreground mass distribution (the lens) deflects light from a distant background source so strongly that the source is observed as multiple images, extended arcs, or, under near-perfect alignment, a complete Einstein ring \citep{schneider1992, treu2010strong}. Recovering the lens and source parameters from such images is an astrophysical inverse problem to which neural density estimation methods have been widely applied \citep{brehmer2019, legin2021simulation, wagner2023images, swierc2024domain}. We simulate observations based on the Euclid VIS instrument configuration \citep{cropper2025euclid}. Each draw is rendered as a noiseless $64\times 64$ image at $0.1$ arcsec/pixel, corrupted with per-pixel noise. The deflector and the source components are together controlled by 17 parameters ($\param \in \mathbb{R}^{17}$). For details on the simulation and training pipeline, see Appendix \ref{sec:lens_app}.

\begin{figure}[ht]
  \begin{center}
    \includegraphics[width=\textwidth]{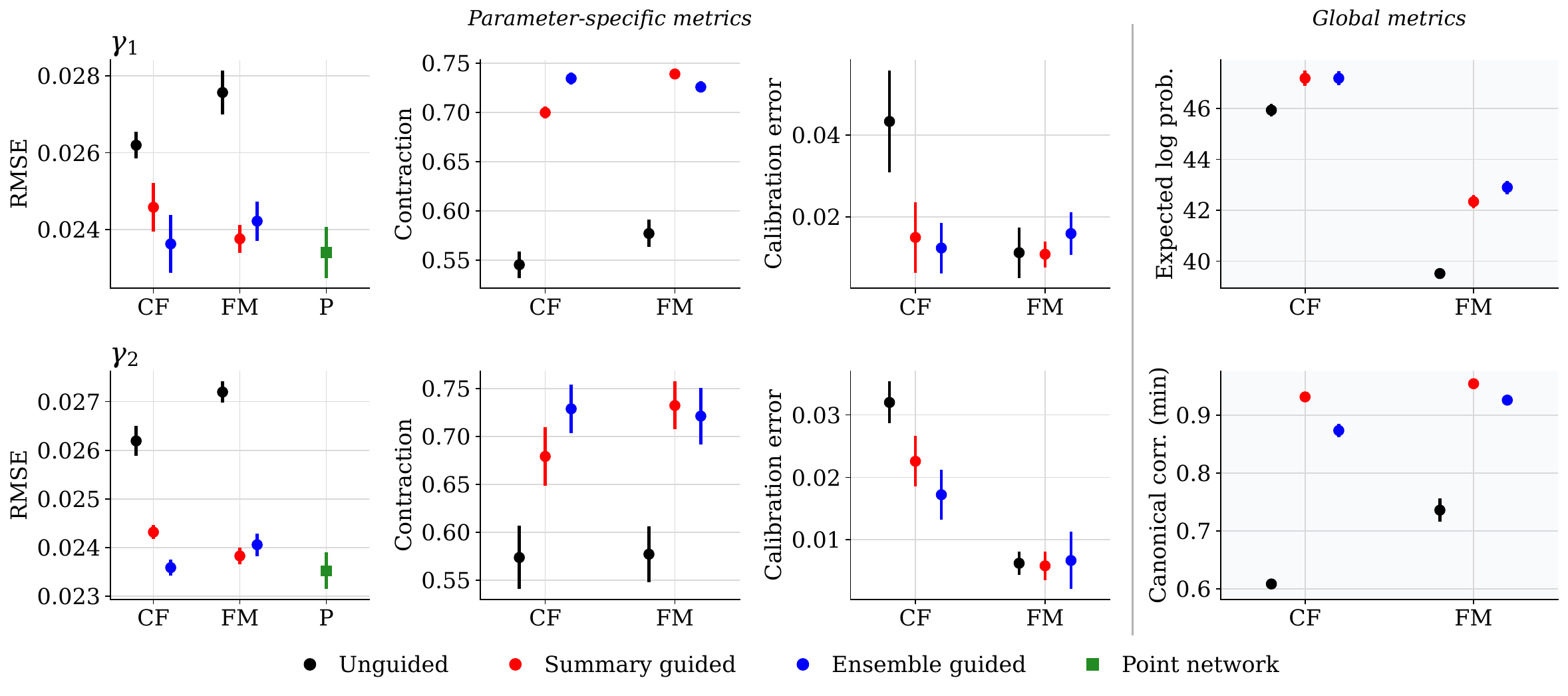}
  \end{center}
  \caption{\emph{Strong gravitational lensing (Experiment 4)}.
    Rows show the two external-shear components, the least identified parameters.
    The right block shows global posterior metrics. Unguided networks suffer from a representation gap as indicated by the CCA diagnostic: $\rho_{\min} = 0.61$ for coupling flow (CF) and $\rho_{\min} = 0.73$ for flow matching (FM).
    Guidance through the summary loss (red) or an ensemble member (blue) clearly improves all metrics and raises $\rho_{\min} \geq 0.87$.
  }
  \label{fig:grav-lens}
\end{figure}

\paragraph{Results.} \autoref{fig:grav-lens} reports the results for the two external-shear parameters ($\gamma_1, \gamma_2$) that are the hardest to recover without guidance along with the global metrics. Both proposed methods improve point-estimate accuracy: for flow matching inference network, both summary- and ensemble-guided training produce RMSE comparable to that of a point network. For coupling flow, ensemble-guided training achieves comparable RMSE as well, summary-guided training with slightly higher RMSE. Posterior contraction improves markedly in all cases and follows the same pattern: coupling flow with summary-guidance slightly lags behind flow matching and coupling flows with ensemble guidance. Crucially, this contraction does not come at the cost of calibration. On the contrary, the calibration error even decreases with training guidance. Moreover, guided training improves inference for the whole joint posterior: both summary- and ensemble-guiding improve expected log probability. The CCA diagnostic further shows that the summary outputs of the guided networks are more informative of the full 17-dimensional posterior than the unguided baseline network.

\subsection{Experiment 5: Social interactions between mice}

We investigate how social interactions among free-ranging mice, represented as a social graph, shape the composition of their gut microbiome. The experimental setup and graph simulator used here are grounded in real-world observational data \citep{raulo_wild_2023, raulo_social_2024}.
The parameters of interest in our simulated setup are the network density of the social graph and the exchange factor, which governs the amount of taxa transferred between two mice upon contact.
In \cite{jedhoff2026}, this setup was used to compare the performance of four graph-suitable summary networks for this inference task, each combined with a coupling flow as the inference network. Results varied strongly across employed summary networks, suggesting the presence of representation gaps. To this end, we employ our guidance losses. More details are in Appendix~\ref{sec:mice_app}.

\begin{figure}
    \centering
    \includegraphics[width=\linewidth]{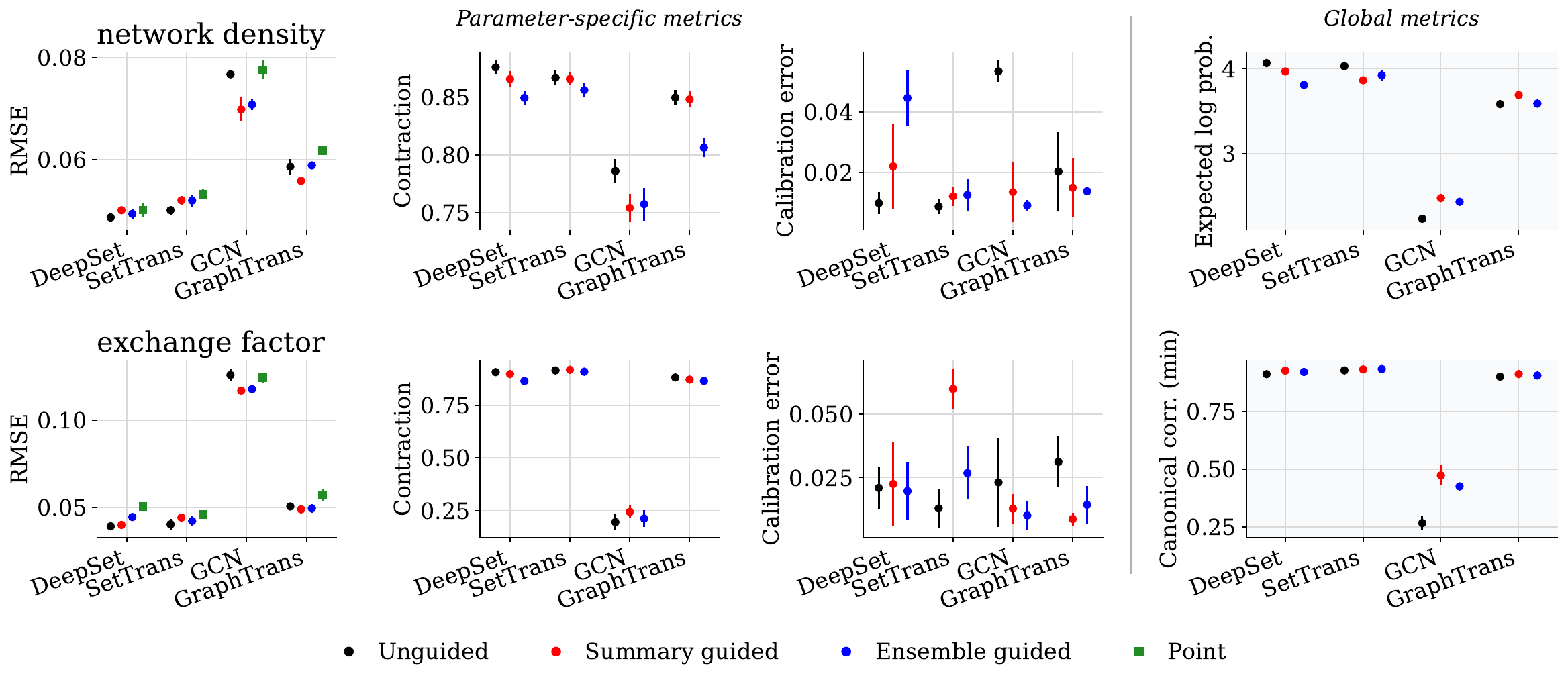}
    \caption{\emph{Social interactions between mice (Experiment 5)}. Parameter-specific metrics for the two parameters network density and exchange factor, and global metrics. Metrics are grouped by the four different summary networks and averaged over three independent test sets of 500 simulations each. Error bars represent standard deviation across the test sets.}
    \label{fig:mice-results}
\end{figure}

\paragraph{Results.} Figure~\ref{fig:mice-results} shows the parameter-specific and global metrics, grouped by the four summary networks. The graph convolutional network (GCN) performs worse than the other three summary networks, consistent with previous findings \citep{jedhoff2026}. For the GCN, summary-guided and ensemble-guided training lead to small improvements, though not enough to match the performance of the other summary networks. For the latter, guided training does not lead to any improvements, which is confirmed by the CCA diagnostic already attaining high values under unguided training in this case. What is more, the point estimation network does not provide better RMSE than the unguided inference network given any of the summary networks. This indicates that little further improvement is possible through guided training and that variations among summary networks are unlikely to stem from a representation gap caused by flawed training dynamics.

\section{Conclusion}

This paper aims to close the representation gap in amortized Bayesian inference that arises when summary networks fail to learn maximally informative statistics. To this end, we introduce auxiliary losses based on posterior point estimates that guide training.
We evaluate two versions, summary-guided and ensemble-guided training, across one toy and four real-world case studies. In all toy settings, both losses clearly speed up convergence and improve recovery. Across three real-world case studies, guided training substantially improves recovery of poorly-estimated parameters, without sacrificing uncertainty calibration. Our proposed diagnostic confirms that these gaps trace back to insufficient summaries. In a fourth real-world case study, guidance yields no consistent improvement---indicating the poor recovery there stems from architectural representation gaps that cannot be fixed by improving training dynamics alone.
This also points to a limitation of our method: If neural point estimation cannot provide better parameter recovery than the unguided inference network, there is no room for our guidance losses to improve performance.
Another limitation is our focus on the posterior mean for guidance. This choice suits our case studies, where the representation gap primarily affected parameter recovery. When the gap instead manifests as poor uncertainty calibration, other point estimates, such as posterior variances or tail quantiles, may be better suited for guidance. Their effectiveness remains an open question for future research.

\newpage

\ifanon\else
\subsection*{Acknowledgements}

This work was partially funded by the U.S. National Science Foundation under Grant No. 2448380, the Deutsche Forschungsgemeinschaft (DFG, German Research Foundation) Project 528702768 as well as DFG Collaborative Research Center 391 (Spatio-Temporal Statistics for the Transition of Energy and Transport) – 520388526.
Furthermore, we also thank the Department of Statistics at TU Dortmund University for providing computing resources and the developers of lenstronomy \citep{birrer2018, birrer2021} for making their gravitational lensing software publicly available.
\fi

\ifarxiv\else
\subsection*{AI use statement}

In this work, we used generative AI tools (Claude) for tasks with recommended disclosure: editing parts of the paper to improve readability and conciseness, and creating software code for summarizing and aggregating case study results, such as figures. We did not use generative AI tools for any tasks with required disclosure, including generating synthetic data, developing theoretical models or conceptual frameworks, formulating mathematical claims, proving mathematical claims, proposing or refining hypotheses, designing research methodology or experiments, implementing methods, or interpreting results. All AI-assisted text was reviewed and edited by the authors for accuracy and to ensure it reflects the authors' intended meaning. All AI-generated code was reviewed and tested for correctness by the authors before use. We take full responsibility for the final content of this work, including all text and code produced with AI assistance.

\subsection*{Reproducibility statement}
All results of our experiments and code that reproduces them will be made publicly available upon acceptance of this paper.
\fi

\bibliography{references}

\appendix
\section{Appendix}

The Appendix contains further details on the experiments (Section \ref{sec:experiments}). We first describe the evaluation metrics used throughout the experiments in more detail, then outline the network architectures and training concepts shared across experiments. We then present each of the five experiments in turn, detailing its simulation setup, any experiment-specific architecture and training choices, and, where applicable, additional results.

\subsection{Metrics}
\label{sec:metrics_app}

Let $m$ index test datasets over which to compute metrics, $j$ index parameters, and $s$ index posterior samples.
The RMSE measures performance of the posterior mean $\bar\param_j^{(m)} = \frac{1}{S}\sum_{s=1}^{S} \hat\param_j^{(m,s)}$ to recover the ground-truth parameter for test dataset $m$:
\begin{align}
    \mathrm{RMSE}_j = \sqrt{\frac{1}{M} \sum_{m=1}^{M} \Big( \bar\param_j^{(m)} - \param_j^{(m)} \Big)^2 } .
\end{align}
For the point-estimate network, where no posterior samples are available, $\bar\param_j^{(m)}$ is directly given by the network's output.

Posterior contraction quantifies how much $\infnet{(\param \given \data)}$ narrows relative to the prior $\prob{\param}$, and is a standard diagnostic in the Bayesian workflow validation \citep{schad2021toward}. With empirical posterior variance $\hat\sigma^{2,(m)}_j$ and prior variance $\sigma^2_{j,\mathrm{prior}}$ estimated empirically from the prior draws $\{\param_j^{(m)}\}_{m=1}^M$, the posterior contraction is
\begin{align}
\mathrm{PC}_j = 1 - \frac{1}{M}\sum_{m=1}^{M} \frac{\hat\sigma^{2,(m)}_j}{\sigma^2_{j,\mathrm{prior}}} ,
\end{align}
with $\mathrm{PC}_j \to 1$ indicating that $\infnet{(\param \given \data)}$ is far more concentrated than the prior, and $\mathrm{PC}_j \to 0$ indicating no gain in information from $\data$.

We assess whether $\infnet{(\param \given \data)}$ is calibrated via simulation-based calibration (SBC) coverage \citep{cook_validation_2006, talts_validating_2020, modrak_simulation-based_2025}. For a grid of $R$ nominal central credible-interval levels $q_r \in [q_{\min}, q_{\max}]$ (with $R=20$ and $[q_{\min}, q_{\max}] = [0.005, 0.995]$; \texttt{BayesFlow} defaults), the empirical coverage is the fraction of test instances for which $\param_j^{(m)}$ falls within the $q_r$-credible interval $\mathrm{CI}_{q_r}(\cdot)$, estimated from the posterior draws:
\begin{align}
    \hat{C}_j(q_r) = \frac{1}{M}\sum_{m=1}^{M} \mathbbm{1}\Big[ \param_j^{(m)} \in \mathrm{CI}_{q_r}\big(\hat\param_j^{(m,1:S)}\big) \Big].
\end{align}
The calibration error aggregates the deviation from nominal coverage across this grid,
\begin{equation}
\mathrm{CE}_j = \operatorname{median}_{r=1,\dots,R} \big| \hat{C}_j(q_r) - q_r \big| ,
\end{equation}
with $\mathrm{CE}_j = 0$ corresponding to perfect calibration.

We also use two global metrics, which are parameter independent.
One of them is the expected log probability, computed as
\begin{align}
    \text{ELP} = \frac{1}{M}\sum_{m=1}^{M} \log\infnet (\param^{(m)} \given \data^{(m)})
\end{align}
This corresponds to the log probability that the trained model assigns to the true parameters $\param^{(m)}$ given the test data conditions $\data^{(m)}$. A high expected log probability is favorable.

As a further diagnostic, we report the canonical correlation analysis (CCA, Section~\ref{sec:canoncial_corr_meth}) between summaries and parameters. The metric indicates how much information about the parameters is retained in the summaries. Against a posterior-mean target, $\rho_{\min} \approx 1$ is desirable.
Against ground-truth parameters, $\rho_{\min}$ is capped by the posterior variance.

\subsection{Properties of the CCA diagnostic}
\label{sec:canoncial_corr_app}

The CCA diagnostic of Section~\ref{sec:canoncial_corr_meth}
can be computed against the ground-truth parameters $\param$, or the posterior mean $\vmu = \Exp{\param \given \data}$.
Here we explain how the interpretation of $\rho_{\min}$ differs.

Split each parameter draw into the posterior mean and the remaining deviation,
\begin{equation}
    \param = \vmu + \vvarepsilon, \qquad \vvarepsilon = \param - \Exp{\param \given \data}.
\end{equation}
The mean $\vmu$ is what the data $\data$ reveal about the first moment of $\param$ and the deviation $\vvarepsilon$ is the posterior uncertainty that remains.
Note that $\vvarepsilon$ is uncorrelated with any function of $\data$.

The summary $\summarynet(\data)$ is such a function, so it co-varies with $\param$ only through $\vmu$:
\begin{equation}
    \Cov(\summary, \param) = \Cov(\summary, \vmu).
\end{equation}
Both targets therefore have the same cross-covariance with the summary and they differ only in their own variance, which for $\param$ additionally contains the posterior uncertainty $\vvarepsilon$.
Two consequences follow.

\paragraph{Ground-truth targets.}
Even when the CCA diagnostic is computed against ground-truth parameters $\param$,
the best any summary can do is to recover $\vmu$ exactly, since the deviation $\vvarepsilon$ cannot be predicted from $\data$.
Hence $\rho_{\min}(\summary,\param) \leq \sqrt{1-v}$, where $v$ is the fraction of prior variance that remains in the posterior, on average, in the least identified parameter direction.
Even a sufficient summary thus scores $\rho_{\min} < 1$ against ground truth, and canonical correlations against $\vmu$ are never lower than against $\param$, as seen in \autoref{fig:crisp-can-cor}.

\paragraph{Posterior-mean targets.}
The canonical correlation $\rho_{\min}(\summary,\vmu)$ is the worst-case correlation between a direction of $\vmu$ and its best linear prediction from $\summary$.
If $\rho_{\min}(\summary,\vmu) \approx 1$, then $\vmu$ is recoverable from $\summary$ in every direction, so conditioning on $\summary$ instead of $\data$ loses almost no information about the posterior mean.
The converse does not hold: a small $\rho_{\min}$ may reflect a non-linear dependence of $\vmu$ on $\summary$ rather than lost information.

\subsection{Network architectures and training}

All experiments are run using \texttt{BayesFlow 2.0.12} \citep{kuhmichel2026} with the JAX backend. %
Unless stated otherwise, networks and training use default hyperparameters. We consider three inference network types: coupling flows, flow matching, and a point network trained with a scoring rule to estimate the posterior mean. For summary networks, we chose architectures suited to each experiment's data.

Beyond architecture, we varied the training regime: \emph{unguided}, in which the inference network trains directly on the summary network's output with no auxiliary signal; \emph{summary guided}, in which a point-estimation loss (mean score) on the summary embedding is combined with the distributional loss via a weighting term; and \emph{ensemble guided}, in which a separate point-estimation network is trained jointly with the distributional inference network, again combined via a weighting term. For the guided regimes, the weight $\lambda$ was either fixed or set automatically, depending on the experiment.

Training was performed either online or offline, depending on the cost of the simulator, using Adam (online) or AdamW (offline) with a cosine-decay learning-rate schedule. Explicit settings for each experiment and its simulation setup are detailed below, if they differ from the default settings.

\subsection{Choosing the auxiliary loss weight}
\label{sec:choose-weight}

To ensure a relevant impact on the training dynamics, the auxiliary guidance loss needs to be of comparable scale to the generative network's loss. We express the guidance loss weight $\lambda$ with respect to normalized magnitudes of both losses on a freshly initialized network with a batch of simulated data, then keeping it fixed throughout training. This proved to be a robust policy that gives comparable results across architectures.
Concretely, we choose $\tilde \lambda$ as a multiplicative scaling factor from which to compute the architecture-specific weight as $\lambda = \tilde \lambda \, |\overline{\score}| / |\overline{\ell}|$. The initial-batch losses of the generative network and the guidance term, respectively, are defined as
\begin{equation}
  \bar{\score} = \frac{1}{B} \sum_{b=1}^{B}
    \score\big(\post{\cdot \given \summary(\data^{(b)})},\, \param^{(b)}\big),
  \qquad
  \bar{\ell} = \frac{1}{B} \sum_{b=1}^{B} \sum_{d=1}^{D}
    \ell\big(\summary_{d}(\data^{(b)}),\, \param^{(b)}_d\big).
\end{equation}
For ensemble guidance, $\summary_d(\data)$ is replaced by $f(\summary(\data))_d$.
In our experiments, we found the guidance to be similarly
effective for any $\tilde\lambda \in [1, 16]$.

\subsection{Experiment 1: Simple location-scale models} \label{sec:location_scale_app}

\paragraph{Simulation.} Every task draws $n=10$ i.i.d. observations per dataset, with all components independent, and differs only in prior and likelihood. Below, $j$ indexes parameter components and $i$ the observations.

\textit{Normal} ($D=64$), whose posterior is available in closed form by conjugacy:
\begin{equation}
    \mu_j \sim \mathcal{N}(0, 1),
    \qquad
    x_{ij} \mid \mu_j \sim \mathcal{N}(\mu_j, 1).
\end{equation}

\textit{Uniform} ($D=32$, i.e.\ $D/2 = 16$ blocks of a center and a half-width), the only task whose prior is not standard normal, since the half-width must stay positive:
\begin{equation}
    c_j \sim \mathcal{N}(0, 1),
    \qquad
    \log h_j \sim \mathcal{N}(0, 0.5^2),
    \qquad
    x_{ij} \mid c_j, h_j \sim \text{Uniform}(c_j - h_j,\; c_j + h_j).
\end{equation}

\textit{Student-t} ($D=16$) with $\nu = 2.5$ degrees of freedom and unit scale:
\begin{equation}
    \mu_j \sim \mathcal{N}(0, 1),
    \qquad
    x_{ij} \mid \mu_j \sim \mu_j + t_{\nu}.
\end{equation}

For \textit{Normal}, the posterior is Gaussian in closed form. \textit{Uniform} and \textit{Student-t} have none, but both factorize over components given the data, so each block's posterior is integrated on a fixed grid over its one- or two-dimensional support.
These references supply the posterior means that the CCA diagnostic correlates against and the expected log probabilities the trajectories are compared to.

\paragraph{Network architectures and training.} Experimental results shown in \autoref{fig:experiment-1} all employ a set transformer summary network with an affine coupling flow.
The summary dimension is twice the parameter dimension ($128$, $64$ and $32$ for \textit{Normal}, \textit{Uniform} and \textit{Student-t}), which leaves half the summary space unconstrained by guidance, as recommended in Section~\ref{sec:guidance_method}.
Optimization uses \texttt{BayesFlow}'s default policy, AdamW under a cosine-decay schedule with warmup and a batch size of $200$.
Each configuration is repeated for three training seeds, which set both the simulated training data and the network initialization.

Training durations are set per simulation budget so that runs are not step-constrained: $3{,}000$, $9{,}000$ and $25{,}000$ gradient steps for \textit{Normal}, $10{,}000$, $90{,}000$ and $90{,}000$ for \textit{Uniform}, and $4{,}500$, $13{,}500$ and $30{,}000$ for \textit{Student-t}, at the $10$k, $30$k and $100$k budgets respectively.
For \textit{Normal}, each budget was additionally trained twice as long to confirm that the reported trajectories have converged rather than been truncated.
For the other tasks training step sensitivity was probed similarly, although with smaller factors, and conclusively.
Since the \textit{Normal} task reaches the overfitting regime at the two smaller simulation budgets, dropout is raised to $0.2$ and $0.1$ at the $10$k and $30$k budgets and left at the default for $100$k; the other two tasks use the defaults throughout.
The effects of increasing dropout were not found to interact with the gap between unguided and guided performance, yet to ensure a conservative estimation of the benefit of our proposed method, we tuned dropout towards optimal performance of the unguided training via a grid scan.

Because all three tasks admit the computation of analytic or near-analytic reference posteriors, the CCA diagnostic here is computed against the exact posterior mean rather than against a reference network's estimate or the true parameter of the simulated samples.

\subsection{Experiment 2: Human decision making} \label{sec:ddm-app}

\paragraph{Simulation.} We simulate from model~1c of \citet{ghaderi-kangavari2023}, a drift-diffusion model in which the trial-wise encoding time is measured by the N200 latency.
Evidence starts at $\beta\alpha$ and evolves as a Wiener process with drift $\delta$ and unit diffusion coefficient until it hits $0$ or the boundary $\alpha$; the choice is $1$ for the upper boundary and $0$ otherwise. The process is simulated with Euler steps of $5$\,ms, and the decision time $T$ is the number of steps times the step size, without a correction for boundary overshoot.
The encoding time on trial $i$ is uniformly distributed with mean $\mu_e$ and standard deviation $s_\tau$, and the N200 latency $z_i$ is a truncated normal centered on it:
\begin{equation}
    \begin{aligned}
        \tau_{e,i} & \sim \text{Uniform}\big(\mu_e - \sqrt{3}\,s_\tau,\; \mu_e + \sqrt{3}\,s_\tau\big) \\
        z_i \mid \tau_{e,i} & \sim \mathcal{N}(\tau_{e,i}, \sigma^2)\ \text{truncated to}\ (0, 0.5) \\
        \text{RT}_i & = T_i + \tau_{e,i} + \tau_m.
    \end{aligned}
\end{equation}
The truncation is implemented by rejection, redrawing $\tau_{e,i}$ together with $z_i$ until $z_i$ lies in the window.
Each dataset consists of 120 independent trials of (RT, choice, $z$), with RT and $z$ in seconds.
The seven free parameters have independent uniform priors:
\begin{equation}
    \begin{aligned}
        \delta & \sim \text{Uniform}(-3, 3) & \alpha & \sim \text{Uniform}(0.5, 2) \\
        \beta & \sim \text{Uniform}(0.1, 0.9) & \mu_e & \sim \text{Uniform}(0.05, 0.40) \\
        \tau_m & \sim \text{Uniform}(0.06, 0.60) & \sigma & \sim \text{Uniform}(0, 0.1) \\
        s_\tau & \sim \text{Uniform}(0, 0.1). &&
    \end{aligned}
\end{equation}
Compared to \citet{ghaderi-kangavari2023}, the prior upper bounds for $\mu_e$, $\tau_m$, $\sigma$ and $s_\tau$ are smaller (paper: $0.6$, $0.8$, $0.3$, $0.3$), the lower truncation bound for $z$ is $0$\,s instead of $0.05$\,s, and $z$ is truncated jointly with $\tau_{e,i}$ rather than conditionally on it.

\begin{figure}
    \centering
    \includegraphics[width=\linewidth]{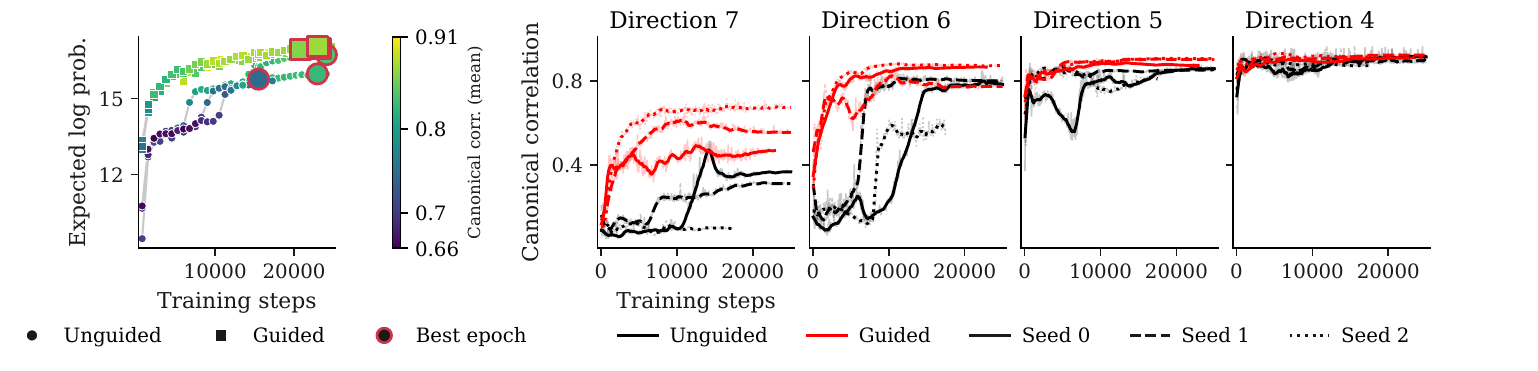}
    \caption{
      Decision making (Experiment 2): Complements \autoref{fig:experiment-2} by showing the trajectories of the last canonical correlations (right) instead of individual parameter's posterior mean accuracy.
    }
    \label{fig:experiment-2-app}
\end{figure}

\paragraph{Networks architectures and Training}
In this experiment we exceptionally use \texttt{BayesFlow} in version 2.0.11, instead of 2.0.12, which includes a modification of the set transformer that is used as the summary network.
The improvements within release 2.0.12 proved to be sufficient to largely fix the \textit{representation gap} with default configurations.
Generally, the failure mode we describe, as any training dynamics issue, is contingent on architecture, initialization, optimizer hyperparameters, and task.
The value of the proposed treatment (\ref{sec:guidance_method}) is that it is \textit{generic} and thus sidesteps the search for a remedy specific to any given combination of these.

\paragraph{Additional results} Employing CCA periodically during training allows us to investigate how linearly decodable information develops.
In \autoref{fig:experiment-2-app} the trajectories of the least informative summary directions are traced, which complements \autoref{fig:experiment-2} from the main text.
The learnt summaries get more informative in distinct steps rather than gradually. Note that consecutive canonical correlations are computed independently from each other and correspond to different coordinate systems.

\subsection{Experiment 3: Eye movements} \label{sec:crisp_app}

\begin{figure}[ht]
    \centering
    \includegraphics[width=0.8\linewidth]{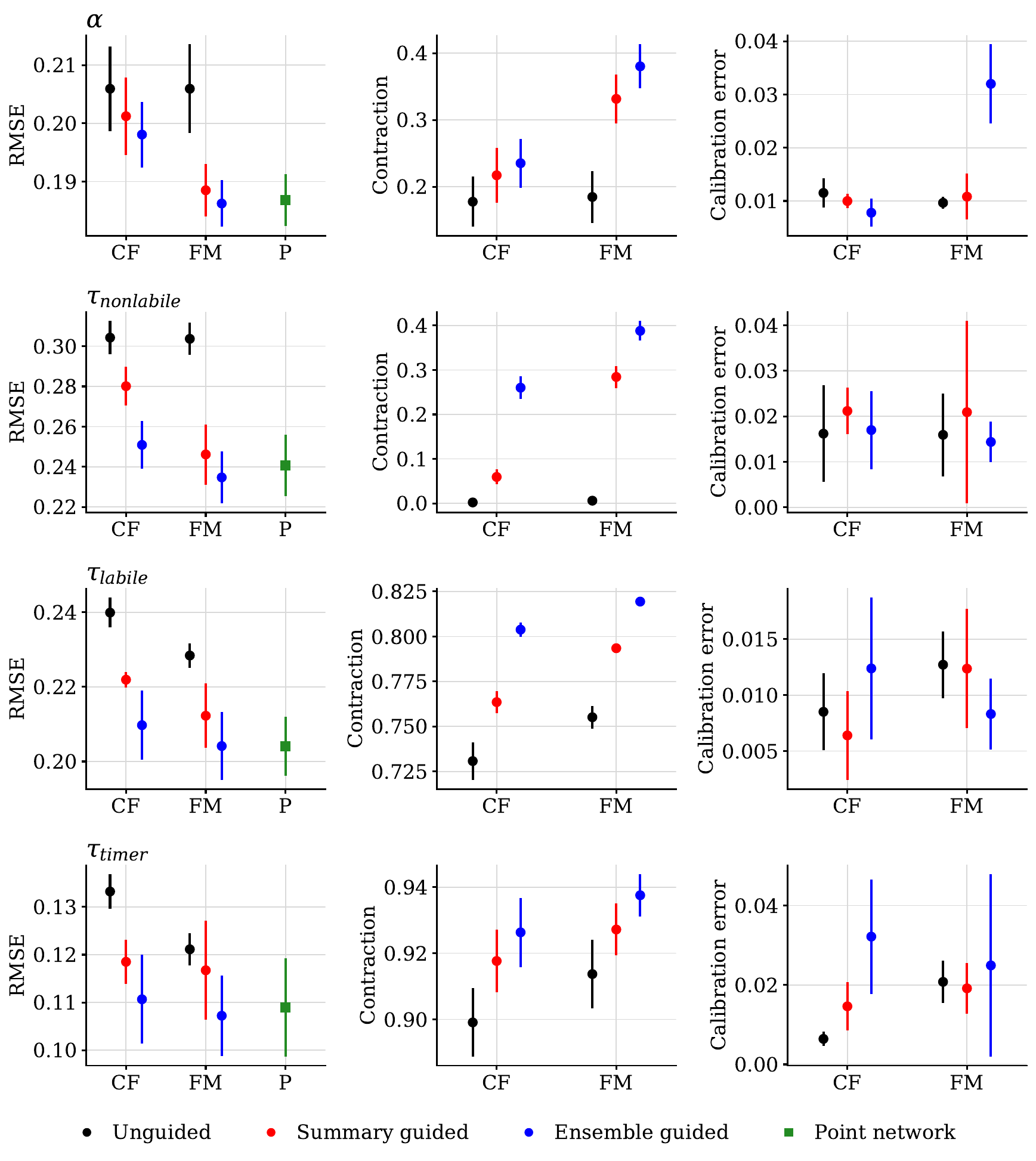}
    \caption{\emph{Experiment 3: Eye Movements}. Parameter-specific metrics for all four parameters of the model.}
    \label{fig:crisp-full}
\end{figure}

\paragraph{Simulation.} We simulate training data from the baseline model described in \citet{walshe2021computational}: Fixation durations are generated as a result of consecutive saccadic programming stages, each an independent discrete random walk with a common threshold $\alpha$ (number of discrete steps to transition to the next stage) but distinct transition rates. Three of the stage rates are inferred ($\tau_\text{timer}$, $\tau_\text{labile}$, and $\tau_\text{nonlabile}$), $\tau_\text{motoric}$ is held fixed at 30ms as per \citet{walshe2021computational}. Following the motoric stage, a saccade is launched with a fixed transition rate $\tau_\text{saccade}=20$ms \citep{walshe2021computational}. Each dataset is composed of 100 independent trials; a trial ends when the maximum number of 40 fixations or maximum trial duration of 25s has been reached, replicating an experimental setting typical for such models \citep{walshe2014asymmetrical}. Priors for the free parameters were set based on the parameter ranges recommended by \citet{walshe2021computational}:
\begin{equation}
    \begin{aligned}
        \alpha & \sim \text{Poisson}(20) \\
        \tau_\text{timer} & \sim \text{Gamma}(6, 35) \\
        \tau_\text{labile} & \sim \text{Gamma}(6, 30) \\
        \tau_\text{nonlabile} & \sim \text{Gamma}(10, 7.5).
    \end{aligned}
\end{equation}
For easier network training, we added a uniform noise to $\alpha$; then, all four parameters were divided by their respective prior mean and log-transformed, making all inference variables unbounded and centered close to zero.

\paragraph{Network architectures and training.}

We trained three inference network types (coupling flow, flow matching, and point network), which were combined with two summary network types (deep set or set transformer), resulting in $3\times2$ architecture combinations.
All summary networks were hierarchical: a bidirectional LSTM network with 16 units first encoded the 40 fixation observation sequences per trial, and its output was pooled across trials by the deep set or set transformer network to produce the final summary output. The summary output had a fixed size of 16.
For the guided training regimes, the weight $\lambda$ was either fixed (10 or 100) or set automatically (see Appendix~\ref{sec:choose-weight}). All models were trained offline on 8192 pre-simulated datasets for 50 epochs of 128 steps each with a batch size of 64.

\begin{figure}[htbp]
    \centering
    \includegraphics[width=\linewidth]{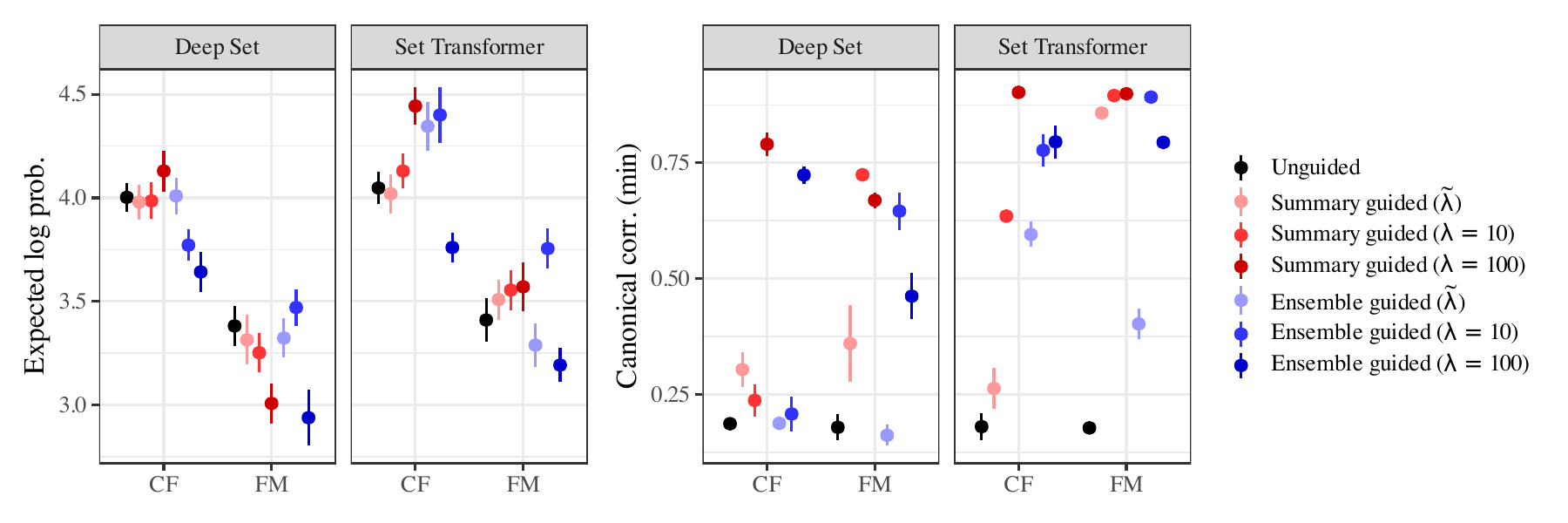}
    \caption{\emph{Experiment 3: Eye Movements}. Expected log probability of the parameters and minimum canonical correlation, across all combinations of inference and summary network architectures, and guidance type and weight. CF=Coupling flow, FM=Flow matching.}
    \label{fig:crisp-overall-global-metrics}
\end{figure}

\begin{figure}[htbp]
    \centering
    \includegraphics[width=\linewidth]{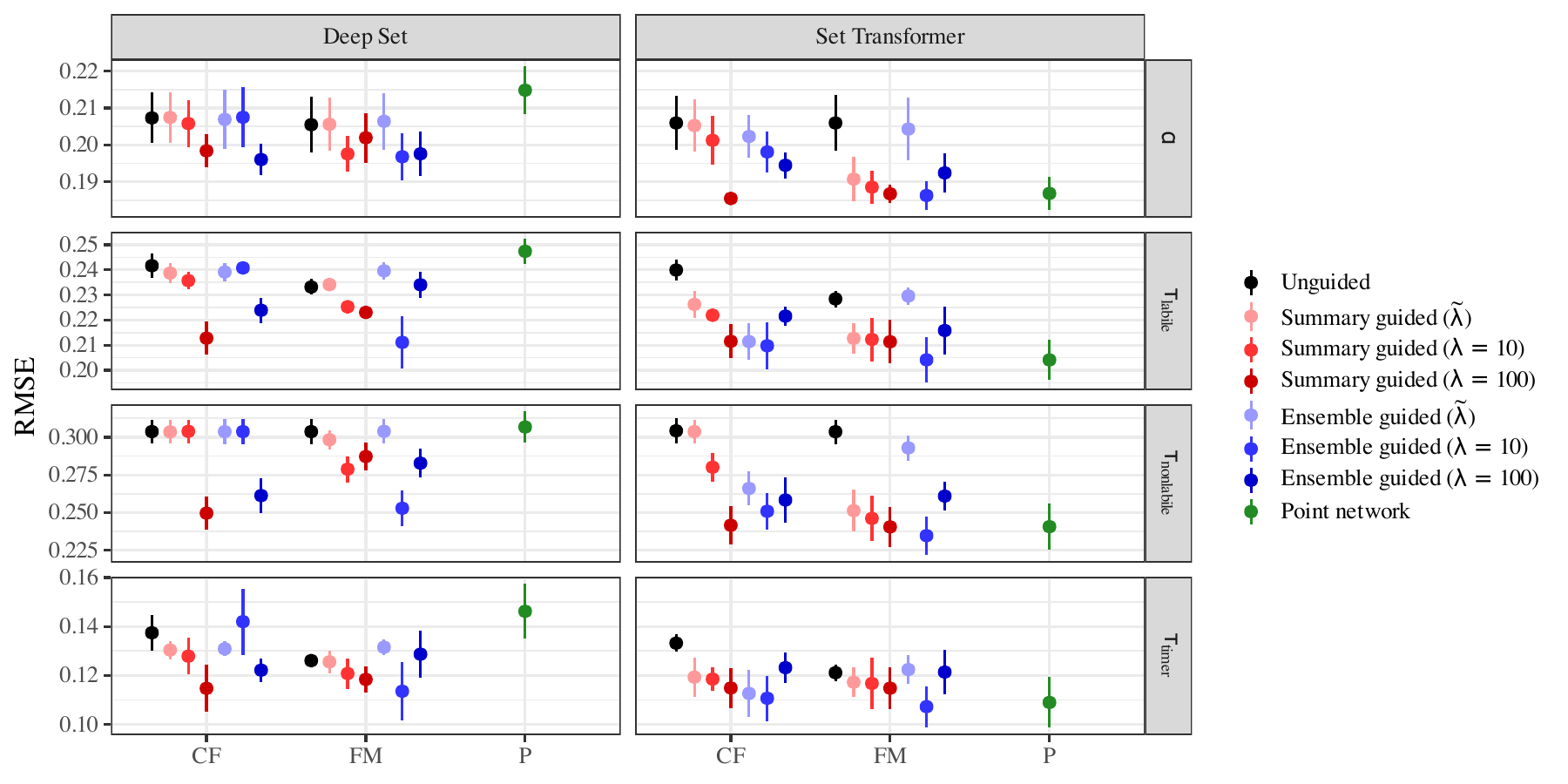}
    \caption{\emph{Experiment 3: Eye Movements}. RMSE of the parameters, across all combinations of inference and summary network architectures, and guidance type and weight. CF=Coupling flow, FM=Flow matching, P=Point network.}
    \label{fig:crisp-overall-rmse}
\end{figure}

\begin{figure}[htbp]
    \centering
    \includegraphics[width=\linewidth]{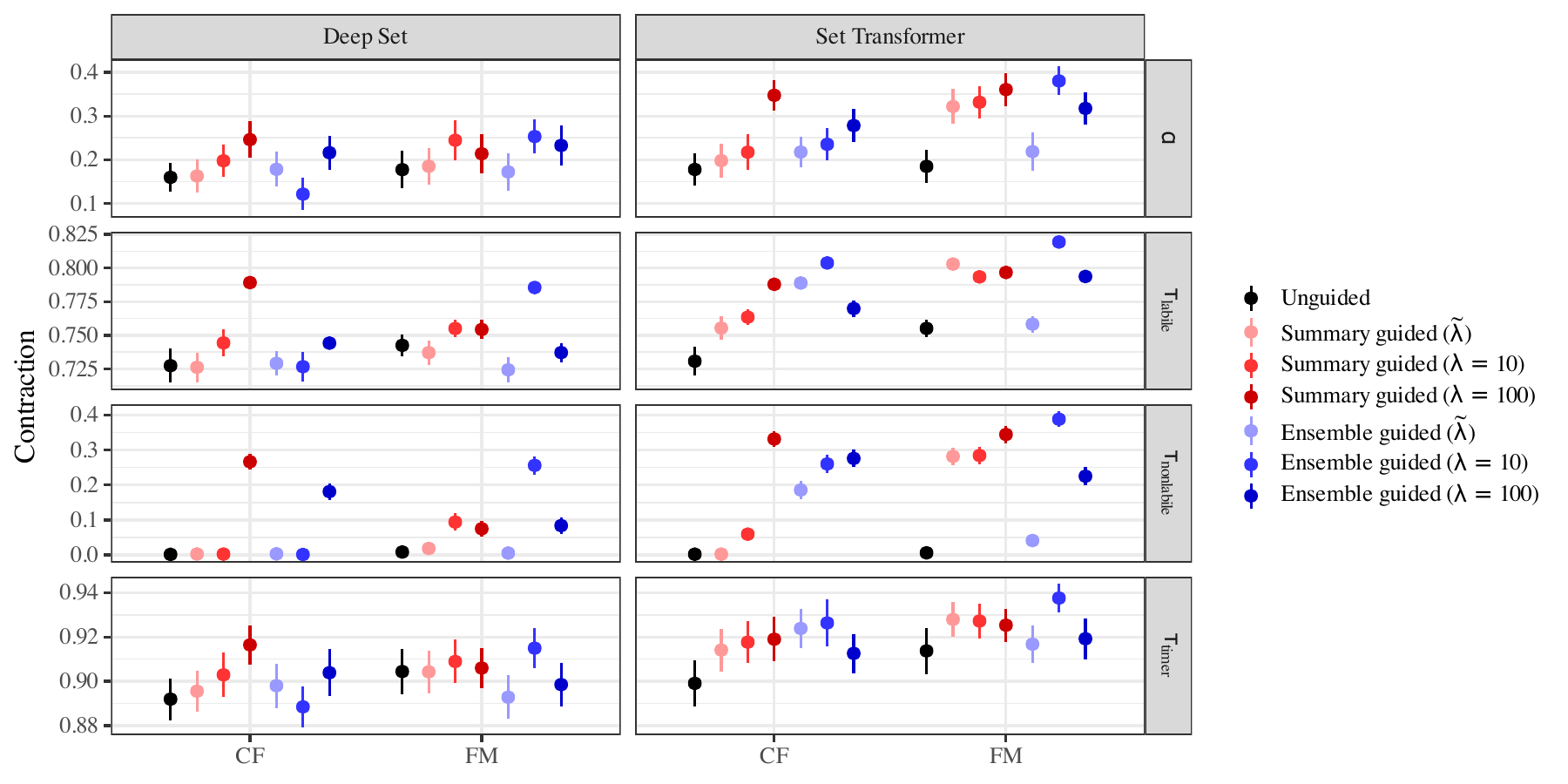}
    \caption{\emph{Experiment 3: Eye Movements}. Posterior contraction of the parameters, across all combinations of inference and summary network architectures, and guidance type and weight. CF=Coupling flow, FM=Flow matching, P=Point network.}
    \label{fig:crisp-overall-contraction}
\end{figure}

\begin{figure}[htbp]
    \centering
    \includegraphics[width=\linewidth]{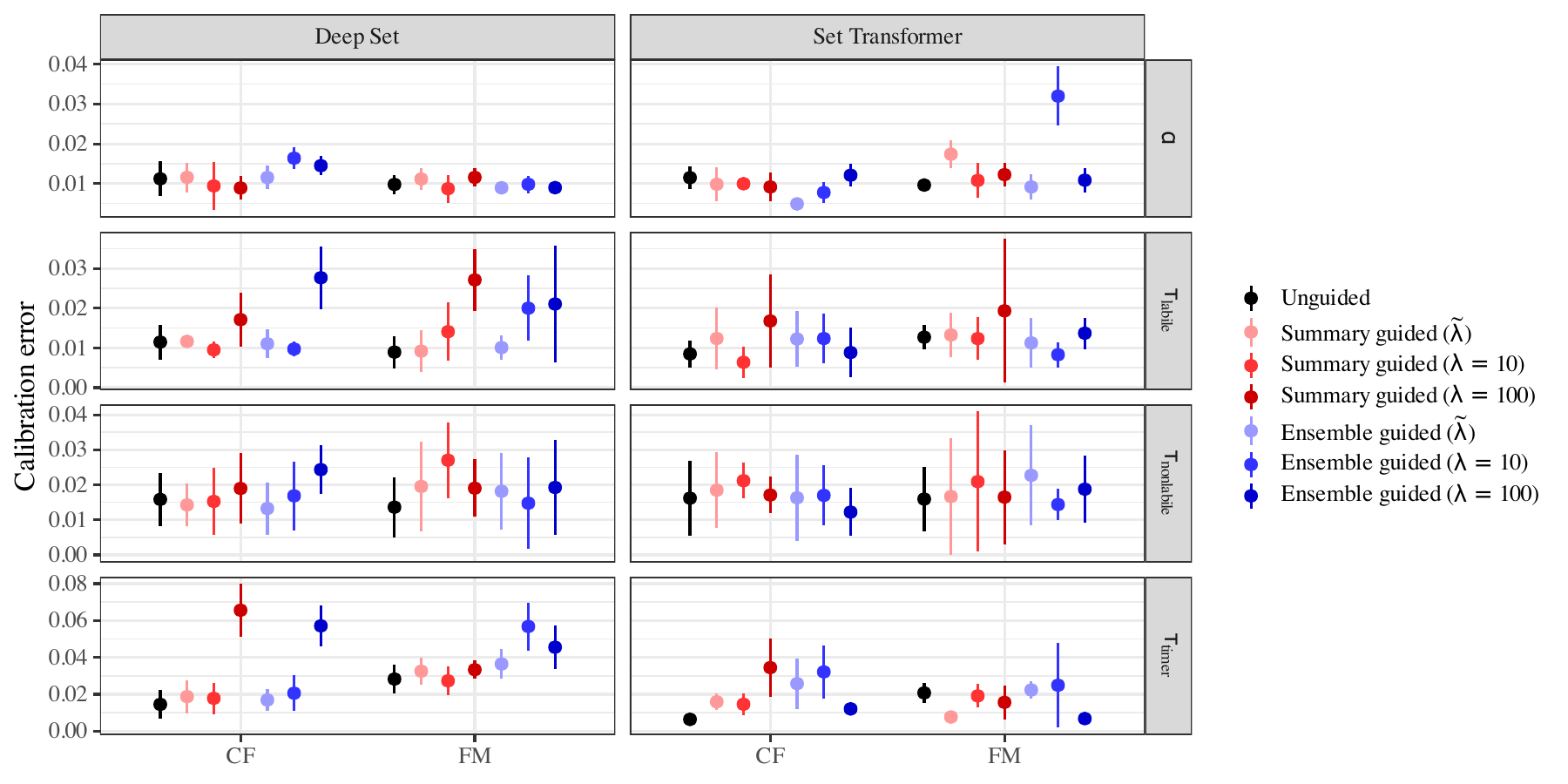}
    \caption{\emph{Experiment 3: Eye Movements}. Calibration error of the parameters, across all combinations of inference and summary network architectures, and guidance type and weight. CF=Coupling flow, FM=Flow matching, P=Point network.}
    \label{fig:crisp-overall-calibration_error}
\end{figure}

\paragraph{Additional results.} The main text reports results only for the two parameters that are hardest to train, using the set transformer summary network and guided training with $\lambda=10$. Figure~\ref{fig:crisp-full} reports all four parameters and shows that guidance helps the other two as well, even though inference for those parameters was already relatively accurate without it. Figures~\ref{fig:crisp-overall-global-metrics}, \ref{fig:crisp-overall-rmse}, \ref{fig:crisp-overall-contraction}, and \ref{fig:crisp-overall-calibration_error} report the full results across all training regimes, both summary network architectures, and all parameters. Overall, guided training tends to match or outperform unguided training, though with considerable variability. In some settings (e.g., $\lambda=100$), guided training improves parameter-specific RMSE but worsens expected joint log probability, suggesting that the point weight overemphasizes the posterior mean at the expense of learning the full posterior within the limited training budget. In others (deep set, coupling flow, $\tilde \lambda=1$ or $\lambda=10$), RMSE and contraction fail to improve for $\tau_\text{nonlabile}$. Together, these results suggest that the ideal point weight is architecture-dependent and may in some cases require careful tuning.

\begin{figure}[htbp]
    \centering
    \includegraphics[width=0.5\linewidth]{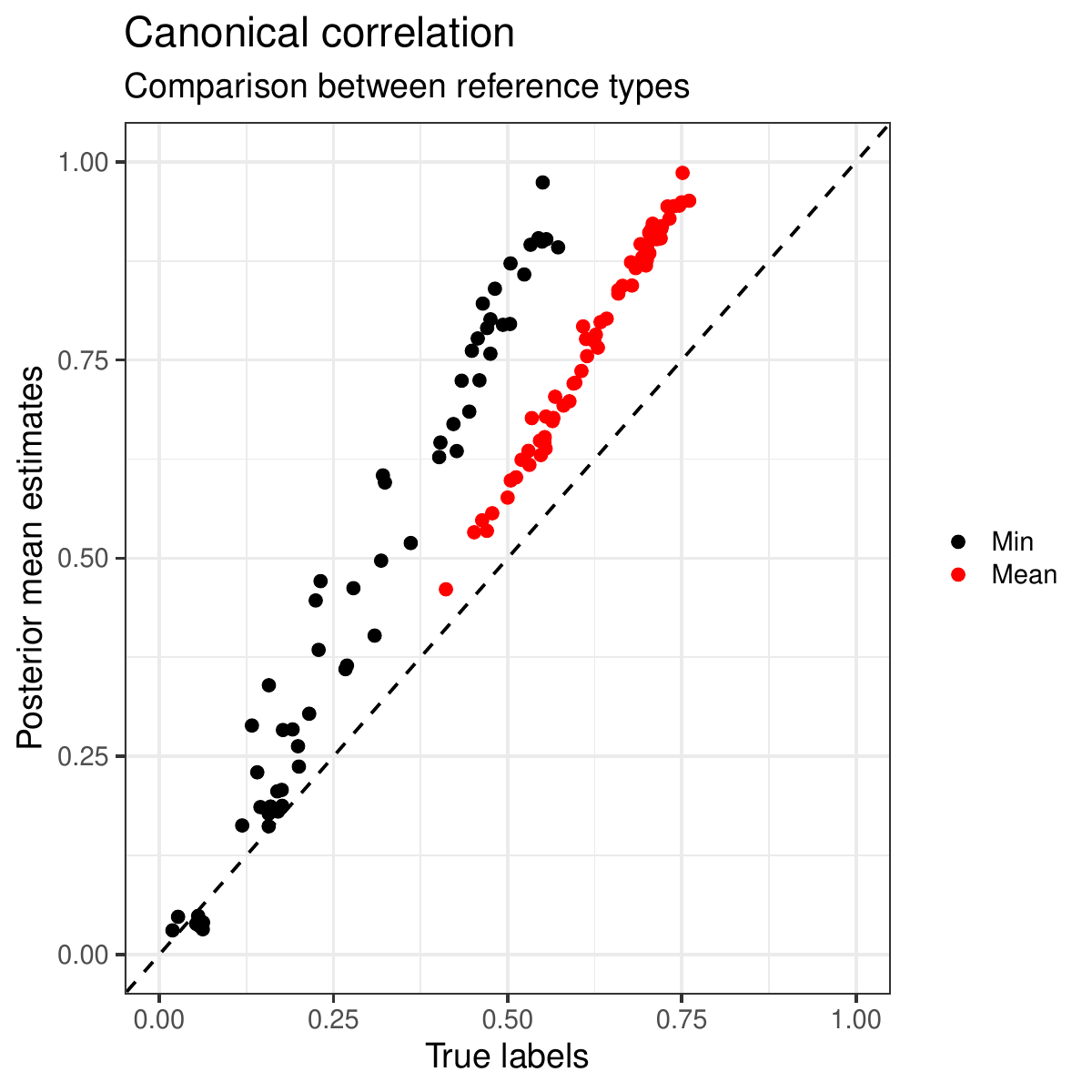}
    \caption{\emph{Experiment 3: Eye Movements}. Minimum and mean canonical correlations computed against the true parameter labels versus against the approximate posterior mean (from the set transformer summary network, 16 output dimensions), across all simulation settings and all three evaluation test sets. Canonical correlations computed against the estimated posterior mean are typically higher than those against the true labels, but the two targets yield strongly correlated diagnostics overall.}
    \label{fig:crisp-can-cor}
\end{figure}

In section~\ref{sec:canoncial_corr_meth}, we explain that canonical correlation may be computed using different targets. In all experiments, that target is the approximate posterior mean from one of the summary networks. In the Eye movement case study, that network is using set transformer with 16 output dimensions as a summary network. In this case study, however, one could also use a canonical correlation with the true parameter labels. Figure~\ref{fig:crisp-can-cor} shows the minimum and mean canonical correlations across all simulation settings and all three evaluation test sets, computed using the true parameter labels as targets vs. using the estimated posterior mean as the target. The figure reveals that the approximate posterior mean typically achieves larger canonical correlations than the true targets; relatively speaking, the two methods correlate with each other rather strongly, suggesting overall agreement between the two options.

\clearpage

\subsection{Experiment 4: Strong gravitational lensing} \label{sec:lens_app}

Gravitational lensing is the deflection of light from a distant source by the gravitational field of an intervening mass distribution (which we refer to as lens). The deflection is described by the lens equation $\beta = \vartheta - \alpha(\vartheta)$, which maps a position $\vartheta$ on the sky to the true source position $\beta$ through the scaled deflection angle $\alpha$ \citep{schneider1992, treu2010strong}. When the projected surface mass density of the deflector exceeds the critical density, this mapping becomes non-injective and the lens equation admits multiple solutions. The source is then observed as several distinct images, extended arcs, or, under near-perfect alignment, a complete Einstein ring. This is the strong lensing regime \citep{treu2010strong}.

\paragraph{Simulation.} We simulate strong lensing images based on the Euclid VIS instrument configuration \citep{cropper2025euclid}. Each simulation draws 17 parameters from their prior distribution (described below) and renders a noiseless image with \texttt{lenstronomy} \citep{birrer2018, birrer2021} after which instrumental noise is added. The mass distribution of the deflector is a singular isothermal ellipsoid (SIE) plus an external shear field. The deflector's own light is an elliptical Sérsic profile and the lensed background source is a second, independent elliptical Sérsic profile. Images are $64\times64$ pixels at $0.1''$/pixel, i.e., a $6.4''\times6.4''$ field of view. Observational settings are taken from the built-in Euclid VIS configuration in \texttt{lenstronomy}. Noise is added as a single per-pixel Gaussian draw with $\sigma^{2}_{ij}=\sigma_\mathrm{bkg}^2 + \mu_{ij}/t_\mathrm{exp}$, where $\mu_{ij}$ is the noiseless model flux and $\sigma_\mathrm{bkg}\approx1.08\times10^{-2}$ counts s$^{-1}$ is the combined sky-plus-read background level and $t_\mathrm{exp}$ is the exposure time. Finally, the noisy image $x$ is passed through $\mathrm{arcsinh}(x/\sigma_\mathrm{bkg})$ transformation which the simulator outputs for better neural network training.

\begin{figure}[h]
    \centering
    \includegraphics[width=\linewidth]{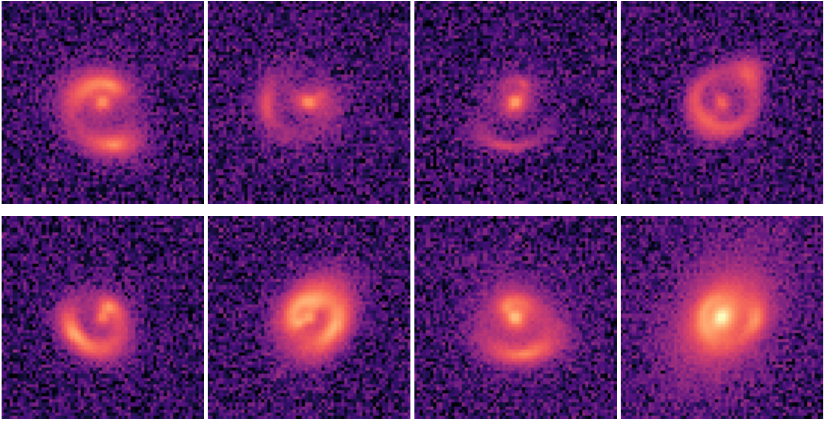}
    \caption{\emph{Experiment 4: Strong gravitational lensing.} Eight independent draws from the prior, each rendered as a single noisy cutout as per the simulation pipeline.}
    \label{fig:lens_img}
\end{figure}

We infer 17 parameters from the observed simulated lensing image. Priors for the deflector are mostly based on the Euclid Q1 data release catalog \citep{euclid2026euclid, walmsley_2025_15025832}. The Einstein radius ($\param_E$) is a $\mathrm{Beta}(2,2)$ rescaled to $[0.5, 1.2]''$. The two mass ellipticity components (${e_1}_m$, ${e_2}_m$)  are independent $\mathcal{N}(0, 0.15^2)$ draws. The deflector's light ellipticity (${e_1}_l$, ${e_2}_l$) components are drawn independently of the mass from a narrower $\mathcal{N}(0, 0.09^2)$. The external shear components ($\gamma_1$, $\gamma_2$) follow $\mathcal{N}(0, 0.04^2)$. The deflector brightness ($\text{mag}_{\text{lens}}$) is $\mathcal{N}(21.3, 1.05^2)$.  Its half-light radius $R_\text{lens}$ is lognormal with median $0.65''$ and range $[0.33, 1.29]''$, slightly tighter than the measured $0.685''$ so that the extended light profile fits inside the $6.4''$ cutout. The deflector Sérsic index $n_\text{lens}$ is lognormal with median $3.7$ and range $[2.70, 5.06]$. The background source being lensed does not have parameters in the released Euclid catalog as it was fitted non-parametrically \citep{euclid2026euclid, walmsley_2025_15025832}; its parameters are therefore set from physical expectation. The source brightness ($\text{mag}_{\text{src}}$) is $\mathcal{N}(23.5, 0.9^2)$ and the size $R_\text{src}$ is lognormal with median $0.25''$ and range $[0.11, 0.55]''$. The source Sérsic index ($n_\text{src}$) is lognormal with median $1.2$ and range $[0.67, 2.16]$, disc-like rather than bulge-dominated. The source ellipticity components (${e_1}_{\text{src}}$, ${e_2}_{\text{src}}$) are sampled from $\mathcal{N}(0, 0.20^2)$. The source position ($x_{\text{src}}, y_{\text{src}}$) follow independent Gaussian distributions around the center $\mathcal{N}(0, 0.20^2)$ while the deflector is fixed at the center. Figure \ref{fig:lens_img} shows eight randomly drawn simulated images as per the simulation pipeline.

\paragraph{Network architecture and training.}
Training is fully online such that each run draws fresh simulations from the simulator at 1563 batches per epoch with batch size 64 for 100 epochs, giving 156,300 gradient steps and $\approx 10^7$ unique simulated images, none of which is ever reused. Validation and test data stay offline and identical across every method: a fixed 4,000-image validation split and three independent 1,000-image test sets are drawn from separate simulator seeds.

Every neural estimator uses the same convolutional summary network with summary dimension of 64, stage widths $(32, 64, 128)$, two residual blocks per stage, and an attention pooling head mapping the $64\times64$ image to a 64-dimensional summary vector. Remaining settings are \texttt{BayesFlow} defaults.
The coupling flow  is six affine coupling layers with random permutations and activation normalisation, whose coupling subnets are three-layer, $256$-wide MLPs which is the sole deviation from the defaults.

\subsection{Experiment 5: Social Interactions between mice}
\label{sec:mice_app}

\paragraph{Simulation.}
We represent a cohort of mice as a weighted interaction network $G=(V,E)$, where nodes $i \in V$ are mice and edge weights $w_{ij} \in [0,1]$ capture expected daily interaction intensity based on spatial proximity, temporal co-occurrence, and/or social association; a global density parameter $\delta$ controls the overall number of ties. Each mouse is initialized with a random subset of microbial taxa. On each discrete daily step, mouse pairs with $w_{ij} > 0$ exchange microbiota in proportion to $w_{ij}$ and an exchange factor $\alpha \in [0,1]$. Repeated over multiple days, this yields gradual convergence of community composition among strongly connected mice, while weakly connected or isolated mice retain more distinct microbiomes; depending on $\delta$ and $\alpha$, the system approaches a steady state after several days.

We simulate a group of 30 mice, each initialized with 5 taxa drawn from a pool of 20. Taxon abundances are encoded in the node-feature matrix $X \in \mathbb{R}^{30 \times 20}$ (rows: mice; columns: taxa; absent taxa as zero, present taxa as relative amounts summing to $100\%$ per mouse). Taxa are removed once their abundance falls below a $0.0001\%$ presence threshold or once they are not exchanged for four consecutive days.

Our goal is to infer the network density $\delta$ and exchange factor $\alpha$ from (i) the interaction network's adjacency matrix and (ii) the final-day microbial composition $X$, under priors $\delta \sim \mathrm{Unif}(0.01,0.5)$ and $\alpha \sim \mathrm{Unif}(0.05,0.5)$. We run the simulator for 5 days.

\paragraph{Network architecture and training.}
For training four different summary network types are considered: deep set, set transformer, graph transformer and a graph convolutional network. The summary output had a fixed size of 8. For the guided training regime, the magnitude normalized weight $\tilde \lambda$ was fixed to 1, see \ref{sec:choose-weight}. Each configuration is trained online for 100 epochs of 100 batches each (batch size 128), validated on 1{,}000 simulations, and evaluated on three independent test sets of 200 simulations each.

\end{document}